\documentclass[10pt,twocolumn,letterpaper]{article}

\usepackage[pagenumbers]{cvpr} 
\usepackage{iftex}
\ifXeTeX
  \usepackage[no-math]{fontspec}
\fi

\graphicspath{{figures/}}

\usepackage{colortbl}
\usepackage{stfloats}
\usepackage{placeins}

\usepackage{algorithm,algpseudocode}
\usepackage[skins,raster]{tcolorbox}
\usetikzlibrary{patterns.meta}

\usepackage{fontawesome5}
\definecolor{cvprblue}{rgb}{0.21,0.49,0.74}
\usepackage[pagebackref,breaklinks,colorlinks,allcolors=cvprblue]{hyperref}

\usepackage{booktabs}
\usepackage{array}
\usepackage{tabularx}
\usepackage{multirow}
\usepackage{colortbl}

\definecolor{NavCoral}{HTML}{FCF1ED}
\definecolor{NavMetricHeader}{HTML}{F5F5F5}

\newcolumntype{N}{>{\raggedleft\arraybackslash}X}

\newcommand{\NavOursName}{\textbf{Dynamic (ours)}}

\usepackage{textcomp} 

\providecolor{MainGray}{HTML}{F4F5F6}
\providecolor{MainAmber}{HTML}{FFF7E6}
\providecolor{MainDeltaBetter}{HTML}{007D88}
\providecolor{MainDeltaWorse}{HTML}{B34B3F}

\providecommand{\MainMetricWithDelta}[4]{%
  \mbox{\makebox[#1][r]{#3}\kern0.3pt%
    \makebox[#2][l]{%
      \raisebox{-0.65ex}{\fontsize{6}{6.4}\selectfont #4}%
    }%
  }%
}

\hypersetup{pdftitle={ReCoNav: On-Demand Compensation for Quantized Vision-and-Language Navigation},
  pdfauthor={Ruizhe Xie, Hao Chen, Yichen Zhang, Yonggang Qi}}

\def\confName{CVPR}
\def\confYear{2027}
\title{ReCoNav: On-Demand Compensation for\\Quantized Vision-and-Language Navigation}
\newcommand{\correspondingmark}{\faEnvelope}
\author{%
  Ruizhe Xie\textsuperscript{1,*}\qquad
  Hao Chen\textsuperscript{2,*}\qquad
  Yichen Zhang\textsuperscript{2}\qquad
  Yonggang Qi\textsuperscript{1,\correspondingmark}\\
  \textsuperscript{1}Beijing University of Posts and Telecommunications\qquad
  \textsuperscript{2}Tsinghua University\\
  {\small \textsuperscript{*}Equal contribution.\qquad
          \correspondingmark\ Corresponding author.}%
}

\begin{document}
\maketitle
\begin{abstract}
Deploying vision-and-language navigation (VLN) models on resource-constrained robots requires balancing inference cost with navigation performance. Low-bit quantization reduces model storage, but numerical approximation can alter actions and subsequent trajectories. Low-rank residual compensation can restore model fidelity, yet applying it at every decision incurs persistent computation. We observe that action disagreement between uncompensated and compensated predictions increases with action uncertainty across navigation backbones, motivating selective fidelity recovery at the decision level. Based on this observation, we introduce \mbox{ReCoNav}, a decision-aware fidelity recovery framework for quantized VLN. ReCoNav augments a fixed low-bit backbone with activation-weighted low-rank corrections, allocating ranks across projections under a fixed storage budget. An uncertainty-guided controller recomputes uncertain action candidates with compensation before execution. When compensated predictions remain uncertain, a stateful continuation mechanism starts the next decision with compensation enabled, avoiding redundant uncompensated inference. Across five model configurations spanning diverse action interfaces, ReCoNav retains near-full-precision navigation success on R2R and RxR in closed-loop simulation. On StreamVLN, it approaches always-on compensation performance while compensating only $39.8\%$ of decisions and reducing mean decision latency from $2.89$ to $2.59$~s on an NVIDIA Jetson Orin NX 16GB. The complete quantized StreamVLN deployment reduces model storage from $14.96$ to $5.58$~GiB and achieves approximately $20\times$ faster inference than its full-precision deployment baseline. ReCoNav also enables on-device inference of SimpleVLN-9B, which exceeds the device's memory at full precision.
Code is available at \href{https://eatsolx.github.io/ReCoNav/}{\mbox{eatsolx.github.io/ReCoNav/}}.
\let\endabstract\par
\end{abstract}

\section{Introduction}
\label{sec:intro}

Vision-and-language navigation (VLN) requires embodied agents to make sequential decisions from visual observations and language instructions while continuously interacting with their environments. Recent video-based navigation models, such as Uni-NaVid and StreamVLN, leverage large multimodal backbones to achieve strong navigation performance across diverse environments~\cite{zhang2025uninavid,wei2025streamvln}. However, repeatedly executing these models within a closed-loop control process imposes substantial memory and computational demands, making their deployment on resource-constrained robots challenging. Post-training quantization offers a practical solution by reducing model storage and inference cost, but the resulting numerical approximation can alter action predictions and, consequently, the trajectories followed by navigation agents.

\begin{figure}[!t]
  \centering
  \includegraphics[width=\columnwidth]{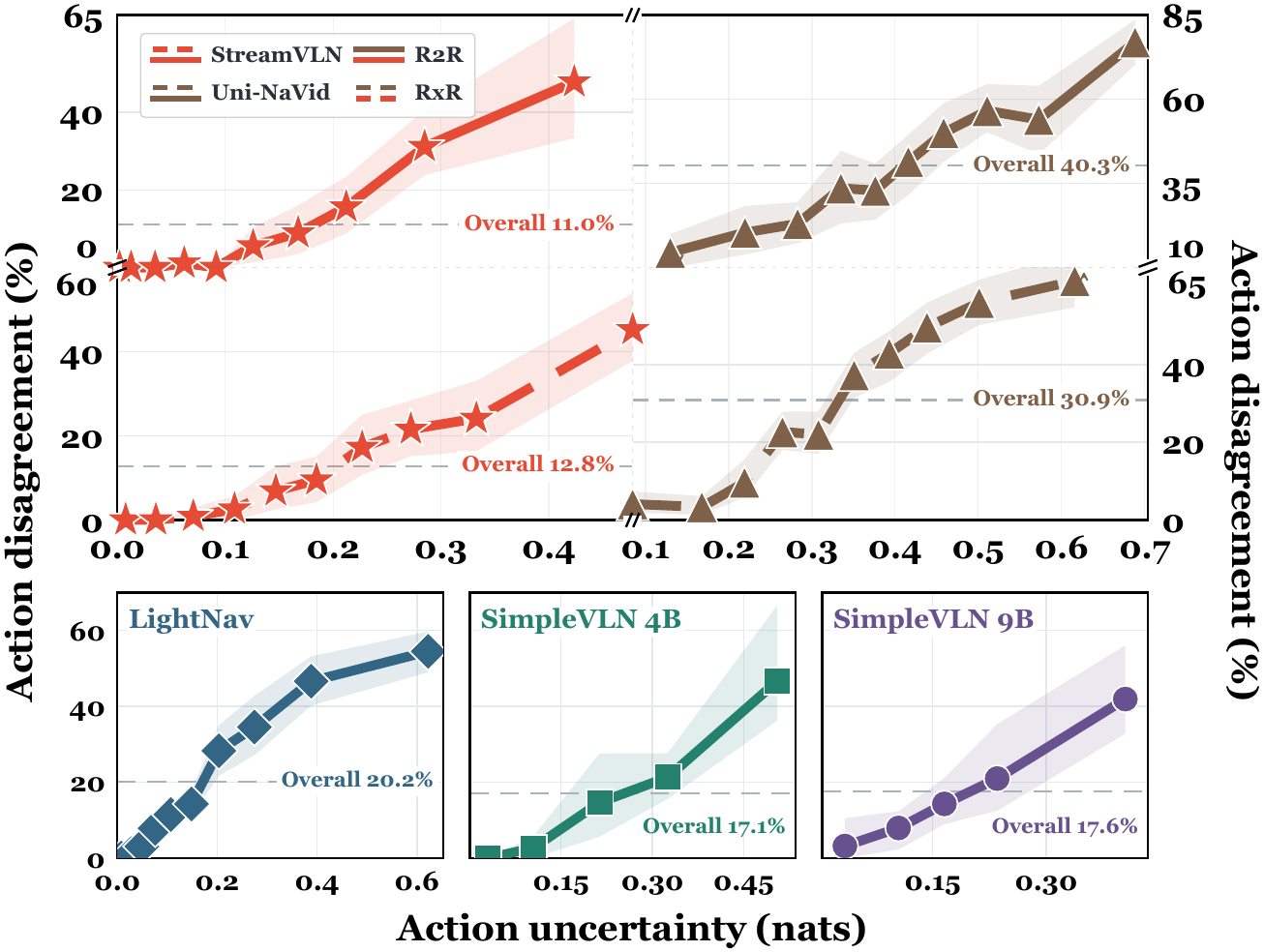}
\caption{\textbf{Action disagreement increases with base uncertainty across five model configurations.}
Disagreement compares uncompensated and compensated predictions from the same decision input. Dashed horizontal lines show the overall disagreement rates. Shading gives path-bootstrap 95\% intervals. Sec.~\ref{app:action-interfaces} details the interface-specific diagnostic.}
  \label{fig:uncertainty}
\end{figure}

Unlike conventional prediction tasks, the impact of numerical approximation in embodied navigation depends on the decisions it affects. A quantization-induced perturbation may leave the selected action unchanged, while another may alter the agent's movement and the observations encountered at subsequent steps. Consequently, minimizing numerical reconstruction error does not necessarily translate into preserving navigation behavior. Existing quantization methods primarily focus on reducing weight or activation errors~\cite{frantar2023gptq,lin2024awq,xiao2023smoothquant,yao2022zeroquant}, while low-rank residual reconstruction further restores the outputs of quantized layers~\cite{zhang2024lqer,zhang2025qera}. Although such compensation improves model fidelity, applying it uniformly throughout navigation incurs additional computation at every decision, regardless of whether the correction is needed. This raises an important question: \emph{Can a quantized navigation model selectively recover its predictive fidelity at decisions sensitive to numerical approximation, rather than maintaining the same compensation cost throughout an entire trajectory?}

To identify decisions sensitive to residual compensation, we compare uncompensated and compensated predictions under identical observations, instructions, and interaction histories. Across navigation models with different action interfaces, disagreement becomes more frequent as the base model's action uncertainty increases (Fig.~\ref{fig:uncertainty}). Action uncertainty, available directly from the base generation, thus provides a practical signal for allocating compensation to decisions whose predicted actions are more likely to change. Whether these changes improve navigation is assessed separately through closed-loop evaluation.

Building on this insight, we introduce \emph{ReCoNav}, a decision-aware fidelity recovery framework for quantized vision-and-language navigation. ReCoNav augments a fixed low-bit backbone with lightweight low-rank residual corrections and activates them on demand according to action uncertainty. To construct these corrections, we employ activation-weighted residual reconstruction and allocate compensation ranks across projections under a fixed storage budget. During navigation, when compensation is disabled, the agent first generates an action candidate using the quantized backbone. If its uncertainty exceeds a calibrated threshold, ReCoNav recomputes the candidate with residual compensation before execution. Beyond this per-decision correction, ReCoNav introduces a stateful continuation mechanism: if the compensated prediction remains uncertain, the subsequent decision begins directly with compensation enabled, avoiding an unnecessary uncompensated generation. Together, these mechanisms enable selective fidelity recovery while limiting the repeated inference cost of compensation.

We evaluate ReCoNav on R2R and RxR~\cite{anderson2018vision,ku2020room} in closed-loop simulation~\cite{krantz2020beyond} across five model configurations, including video-based navigation models, image-grid action prediction, and text-based action generation. ReCoNav consistently achieves navigation success rates close to full-precision references while substantially reducing model storage. On StreamVLN and Uni-NaVid, uncertainty-guided scheduling approaches the performance of always-on compensation while applying corrections to fewer decisions. On an NVIDIA Jetson Orin NX 16GB, ReCoNav further reduces mean decision latency relative to always-on compensation on the same quantized backbone. The complete quantized StreamVLN deployment achieves approximately $20\times$ faster inference than its full-precision baseline, while ReCoNav also enables a 9B-parameter navigation model to operate within the device's memory budget. These results demonstrate the practical value of allocating numerical compensation according to the evolving decisions of embodied agents.

To this end, our main contributions can be summarized as follows: (i) We identify a consistent relationship between action uncertainty and disagreement between base and compensated predictions across different navigation backbones, motivating selective compensation at the decision level. (ii) We introduce ReCoNav, which integrates activation-weighted low-rank reconstruction with uncertainty-guided decision recomputation and temporal compensation continuation to recover predictive fidelity while minimizing redundant computation. (iii) We demonstrate near-full-precision navigation performance across five model configurations on R2R and RxR, alongside reduced memory requirements and inference latency on resource-constrained hardware.

\section{Related Work}
\label{sec:related}

\paragraph{Vision-and-language navigation.}
R2R and RxR benchmark an agent's ability to follow natural-language instructions in visually grounded environments~\cite{anderson2018vision,ku2020room}. VLN-CE extends this task to continuous environments, where agents navigate by executing low-level actions~\cite{krantz2020beyond}. HAMT encodes the history of panoramic observations, while DUET builds an online topological map to connect global planning with local decisions~\cite{chen2021hamt,chen2022duet}. Among video-based models, NaVid predicts actions directly from monocular video and language instructions, Uni-NaVid extends this formulation to multiple embodied tasks, and StreamVLN combines long-term context with recent frames for action generation~\cite{zhang2024navid,zhang2025uninavid,wei2025streamvln}. Navigation models also differ in how they represent actions: LightNav-0 predicts spatial points in the image, while SimpleVLN generates text commands that can be parsed into primitive actions~\cite{wang2026lightnav,hong2026simplevln}. We evaluate ReCoNav primarily on Uni-NaVid and StreamVLN and use LightNav-0 and SimpleVLN to assess transfer across action interfaces.

\paragraph{Post-training quantization and residual reconstruction.}
Post-training quantization uses calibration data to limit the error introduced by low-precision representations. GPTQ uses approximate second-order information to quantize weights layer by layer, while AWQ uses activation statistics to rescale channels and protect salient weights~\cite{frantar2023gptq,lin2024awq}. OmniQuant learns weight clipping and equivalent transformations through blockwise reconstruction~\cite{shao2024omniquant}. To reduce the error that remains after quantization, LQER and QERA use activation statistics to construct low-rank corrections for layer outputs~\cite{zhang2024lqer,zhang2025qera}. Corrections can also adapt to the input: DecDEC retrieves residuals for salient channels during decoding, while SPEAR modulates low-rank corrections according to individual token representations~\cite{park2025decdec,liu2026spear}. ReCoNav controls residual compensation at the level of a complete navigation decision.

\paragraph{Task-aware quantization for embodied control.}
Quantization methods for vision-language-action models account for how numerical errors affect control. QuantVLA identifies sensitive operations in the diffusion action head and calibrates the scales of attention logits and residual updates~\cite{zhang2026quantvla}. QVLA allocates bit widths according to how quantizing individual channels changes the predicted action~\cite{xu2026qvla}. DyQ-VLA exploits the different ways translation and rotation errors evolve during execution: it adjusts activation precision over time and compensates for accumulated rotation errors~\cite{zheng2026dyqvla}. ReCoNav keeps the quantized backbone and low-rank correction factors fixed, and uses action uncertainty to control recomputation before execution and compensation continuation at the next navigation decision.
\section{Method}
\label{sec:vlnq-method}

\begin{figure*}[!t]
  \centering
  \includegraphics[width=\textwidth]{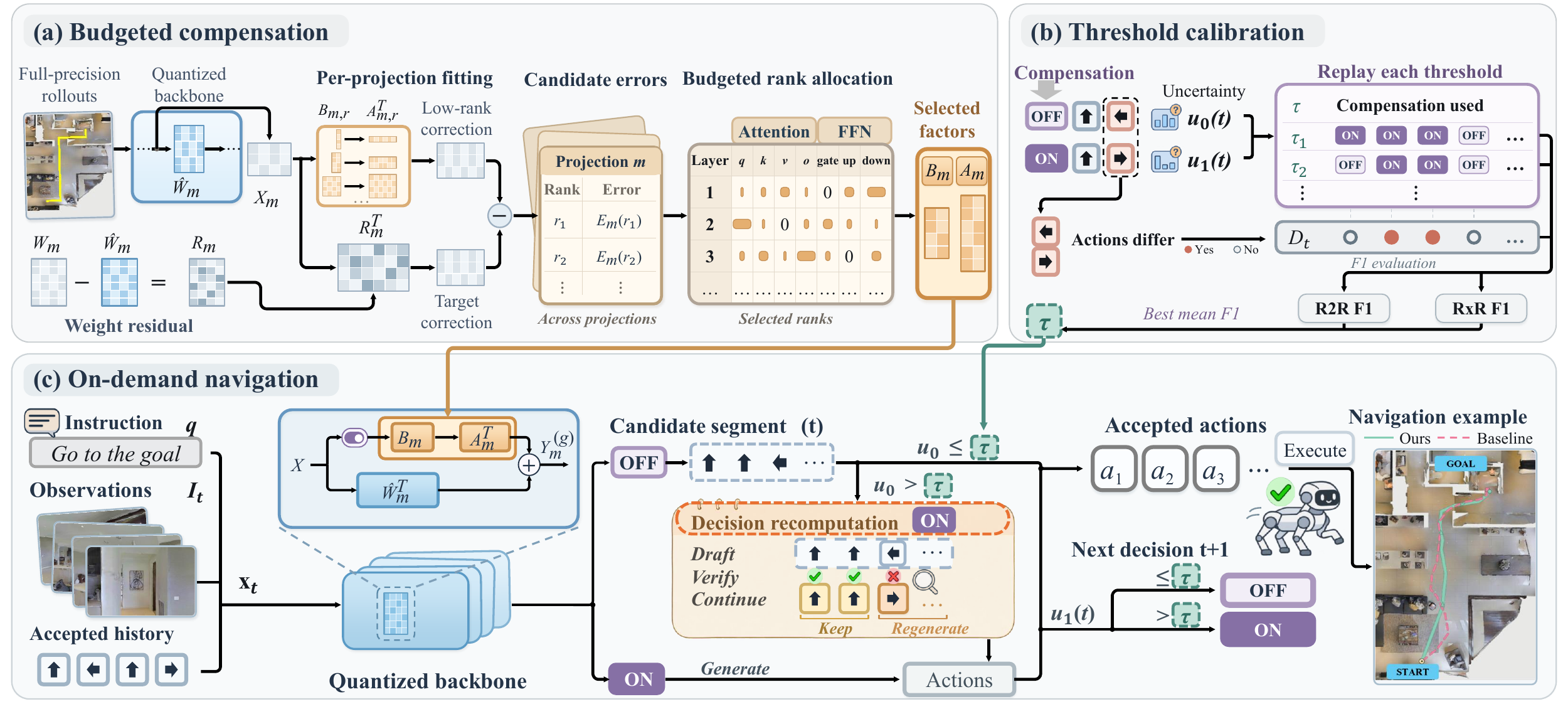}
  \caption{ReCoNav overview: (a) activation-weighted correction with budgeted ranks; (b) per-model threshold calibration from paired predictions; (c) shared-threshold control of recomputation and compensation continuation. Trajectories are schematic.}
  \label{fig:overview}
\end{figure*}

ReCoNav executes low-rank compensation on demand over a fixed quantized backbone. As outlined in~\cref{fig:overview}, offline preparation consists of activation-weighted reconstruction with budgeted rank allocation (a) and threshold calibration from paired predictions (b). During navigation, the controller uses the resulting factors and threshold to schedule compensation (c).

\subsection{Quantized Backbone and Compensation}
\label{sec:vlnq-setup}

At decision $t$, the agent receives $x_t=(I_t,q,h_t)$, comprising visual observations $I_t$, a language instruction $q$, and the accepted interaction history $h_t$. It generates an action segment, executes it, and obtains new observations for the next decision.

We quantize attention and feed-forward projections in the language backbone using packed low-bit formats in llama.cpp~\cite{gerganov2024llamacpp}, retaining the original precision of the visual encoder, multimodal projector, token embeddings, and output head. The primary backbones use the \texttt{IQ3\_XS} preset; model-specific formats and computation precision are specified in Sec.~\ref{app:setup}. Quantized weights remain fixed throughout navigation. Each quantized projection has its own low-rank compensation branch, and all branches share a switch $g\in\{0,1\}$ within a generation.

For projection $m$, let $W_m$ be the original weight matrix and $\widehat W_m$ the matrix represented by the quantized weights after dequantization. We approximate the weight residual $R_m=W_m-\widehat W_m$ with two low-rank factors $A_mB_m^{\mathsf T}$, as illustrated in~\cref{fig:overview}(a). Given input activations $X$, the projection produces
\begin{equation}
Y_m^{(g)}=X\widehat W_m^{\mathsf T}+g(XB_m)A_m^{\mathsf T}.
\label{eq:vlnq-projection}
\end{equation}
Here, $W_m\in\mathbb R^{d_{\mathrm{out},m}\times d_{\mathrm{in},m}}$, $A_m\in\mathbb R^{d_{\mathrm{out},m}\times r_m}$, and $B_m\in\mathbb R^{d_{\mathrm{in},m}\times r_m}$. We omit the unchanged bias. With $g=0$, the layer executes the base quantized projection; with $g=1$, two low-rank matrix multiplications add the residual correction.

\subsection{Offline Reconstruction and Capacity Allocation}
\label{sec:vlnq-reconstruction}

We fit activation-weighted low-rank corrections~\cite{zhang2024lqer,zhang2025qera} using input activations $X_m$ recorded from the quantized backbone on full-precision-driven training trajectories. Holding these inputs fixed, we approximate the output residual $X_mR_m^{\mathsf T}$. For candidate rank $r$, the reconstruction error is
\begin{equation}
\mathcal E_m(r)=\left\|X_mR_m^{\mathsf T}-(X_mB_{m,r})A_{m,r}^{\mathsf T}\right\|_F^2.
\label{eq:vlnq-reconstruction}
\end{equation}
We obtain ordered factors from an activation-subspace decomposition and use their first $r$ columns for candidate rank $r$. Errors are evaluated after conversion to deployment precision.

Let $\mathcal P$ denote eligible projections and $\mathcal R_m\subseteq\{0,16,32,64,128\}$ the ranks supported by projection $m$. With 16-bit factors, we allocate ranks under byte budget $M$:
\begin{equation}
\begin{aligned}
\min_{\{r_m\in\mathcal R_m\}_{m\in\mathcal P}}\quad&\sum_{m\in\mathcal P}\mathcal E_m(r_m),\\
\mathrm{s.t.}\quad&\sum_{m\in\mathcal P}2r_m(d_{\mathrm{in},m}+d_{\mathrm{out},m})\le M,\\
&\frac{1}{|\mathcal P|}\sum_{m\in\mathcal P}r_m\le64.
\end{aligned}
\label{eq:vlnq-rank-allocation}
\end{equation}
The budget $M$ equals the 16-bit factor storage of a uniform rank-$64$ configuration; rank $0$ disables a branch. We approximately solve this allocation using multiple-choice knapsack dynamic programming, tightening the byte budget when needed to satisfy the mean-rank cap.

For each model configuration and dataset, we fit factors and allocate ranks using five training trajectories, and evaluate reconstruction on three additional held-out training trajectories. Factors and ranks are frozen before unseen-split evaluation. The activation-collection protocol, factorization, and rank-allocation solver are detailed in Sec.~\ref{app:compensation}.

\subsection{Uncertainty-Guided Compensation}
\label{sec:vlnq-scheduling}

\paragraph{Action uncertainty.}
The controller reads probability scores already available during action generation. For Uni-NaVid and StreamVLN, let $\mathcal A$ be the agent's four action tokens and $\mathcal J_t^g$ the valid scored positions under compensation state $g$. At position $j$, the probability $p_{t,j}^g$ of the generated action is recomputed by a softmax over $\mathcal A$. We define action-segment uncertainty as
\begin{equation}
u_g(t)=-\frac{1}{|\mathcal J_t^g|}\sum_{j\in\mathcal J_t^g}\log p_{t,j}^g.
\label{eq:vlnq-uncertainty}
\end{equation}
A larger value indicates lower confidence. Calibration and online execution use identical scoring rules, restricting uncertainty to the action-bearing content of each interface: discrete actions, spatial pointers, or parsed command fields (Sec.~\ref{app:action-interfaces}).

\paragraph{Uncertainty and action disagreement.}
We generate paired outputs without and with compensation from the same observations, instruction, and accepted history along training paths. These paired predictions provide the uncertainty scores and action-disagreement labels used for threshold calibration in~\cref{fig:overview}(b). For the two discrete-action backbones, let $\tilde a_t^{(0)}$ and $\tilde a_t^{(1)}$ denote their ordered action prefixes. We define
\begin{equation}
D_t=\mathbf1\!\left\{\tilde a_t^{(0)}\ne\tilde a_t^{(1)}\right\}.
\label{eq:vlnq-disagreement}
\end{equation}
For the other interfaces, $D_t$ compares spatial pointers or canonical primitive-action sequences. We evaluate navigation outcomes separately in closed-loop rollouts.

\paragraph{Current-decision recomputation.}
For the controller in~\cref{fig:overview}(c), let $s_t\in\{0,1\}$ denote the compensation state at the start of decision $t$, and $g_t$ the state used for its accepted action, with $s_1=0$. With compensation off, the controller triggers a rerun only for a valid base score above threshold:
\begin{equation}
 g_t=\begin{cases}
 1,& s_t=0,\ \operatorname{valid}(u_0(t)),\ u_0(t)>\tau,\\
 s_t,& \text{otherwise}.
 \end{cases}
\label{eq:vlnq-accept}\end{equation}
Each decision is rerun at most once, and only the accepted action is executed.

For a compensated rerun, we restore the pre-decision language state and reuse visual features from the uncompensated pass, avoiding repeated visual encoding and visual-history updates. To accelerate the rerun, the compensated model verifies the uncompensated output tokens as a draft in batches, retains the matching prefix, and resumes autoregressive decoding at the first mismatch. Sec.~\ref{app:implementation} describes cache restoration and prefix verification.

\paragraph{Continuation and exit.}
A valid compensated score controls continuation. If the score is unavailable, the controller retains the accepted state:
\begin{equation}
 s_{t+1}=\begin{cases}
 \mathbf{1}\{u_1(t)>\tau\}, & g_t=1,\ \operatorname{valid}(u_1(t)),\\
 g_t,& \text{otherwise}.
 \end{cases}
\label{eq:vlnq-next}\end{equation}
Algorithm~\ref{alg:app-compensation} gives the full control flow.

\paragraph{Shared-threshold calibration.}
\label{sec:vlnq-calibration}
For each model configuration $b$, we replay the complete controller on paired training records for each candidate threshold (\cref{fig:overview}(b)).
We compute disagreement F1 using $D_t=1$ as the positive label and $g_t=1$ as the positive prediction, including reruns and decisions that start compensated.
The shared threshold maximizes the mean score across datasets:
\begin{equation}
\tau_b^*=\operatorname*{arg\,max}_{\tau\in\mathcal T_b}
\frac{F_{1,b}^{\mathrm{R2R}}(\tau)+F_{1,b}^{\mathrm{RxR}}(\tau)}{2}.
\label{eq:vlnq-calibration}
\end{equation}
The selected $\tau_b^*$ is shared across R2R and RxR and governs both triggering and exit. Threshold candidates, paired rollouts, and path-grouped cross-validation follow the protocol in Sec.~\ref{app:calibration}.

\section{Experiments}
\label{sec:experiments}
\begingroup
\definecolor{MainDeltaBetter}{HTML}{007D88}
\definecolor{MainDeltaWorse}{HTML}{B34B3F}
\begin{table*}[t]
\centering
\caption{\textbf{Navigation performance and deployment cost across five model configurations.}
Navigation metrics are measured on the R2R and RxR unseen validation splits; SR, SPL, and OS are percentages, and NE is in meters. Size denotes complete-model storage (GiB), and latency is measured in seconds per decision on Jetson Orin NX 16GB. Subscripts give absolute differences relative to ReCoNav within each model group: arrows indicate numerical direction and teal/red indicates better/worse performance. OOM denotes insufficient device memory. The GPTQ--ReCoNav latency comparison uses successful RxR tasks (Sec.~\ref{app:setup}).}
\label{tab:main}
\begingroup\small
\setlength{\tabcolsep}{2.4pt}
\setlength{\aboverulesep}{0pt}\setlength{\belowrulesep}{0pt}
\renewcommand{\arraystretch}{1.18}
\renewcommand{\tabularxcolumn}[1]{m{#1}}

\begin{tabularx}{\linewidth}{>{\raggedright\arraybackslash}X*{9}{r}}
\toprule
\multirow{2}{*}[-0.6pt]{\textbf{Configuration}} & \multicolumn{4}{c}{\textbf{R2R}} & \multicolumn{3}{c}{\textbf{RxR}} & \multicolumn{2}{c}{\textbf{Deployment}} \\
\cmidrule(l{\tabcolsep}r{\tabcolsep}){2-5}\cmidrule(l{\tabcolsep}r{\tabcolsep}){6-8}\cmidrule(l{\tabcolsep}r{\tabcolsep}){9-10}
 & \multicolumn{1}{c}{SR $\uparrow$} & \multicolumn{1}{c}{SPL $\uparrow$} & \multicolumn{1}{c}{NE $\downarrow$} & \multicolumn{1}{c}{OS $\uparrow$} & \multicolumn{1}{c}{SR $\uparrow$} & \multicolumn{1}{c}{SPL $\uparrow$} & \multicolumn{1}{c}{NE $\downarrow$} & \multicolumn{1}{c}{Size $\downarrow$} & \multicolumn{1}{c}{Latency $\downarrow$} \\
\midrule
\rowcolor{MainGray}\multicolumn{10}{l}{\cellcolor{MainGray}\rule{0pt}{2.6ex}\underline{\textit{StreamVLN-7B}}} \\
\hspace*{9pt}Full precision & \MainMetricWithDelta{20.5pt}{18pt}{56.44}{\textcolor{MainDeltaBetter}{\textuparrow\kern0.15pt 0.11}} & \MainMetricWithDelta{20.5pt}{18pt}{50.39}{\textcolor{MainDeltaBetter}{\textuparrow\kern0.15pt 0.25}} & \MainMetricWithDelta{16pt}{15pt}{4.90}{\textcolor{MainDeltaBetter}{\textdownarrow\kern0.15pt 0.05}} & \MainMetricWithDelta{20.5pt}{18pt}{63.64}{\textcolor{MainDeltaBetter}{\textuparrow\kern0.15pt 0.13}} & \MainMetricWithDelta{20.5pt}{18pt}{54.57}{\textcolor{MainDeltaBetter}{\textuparrow\kern0.15pt 0.25}} & \MainMetricWithDelta{20.5pt}{18pt}{45.52}{\textcolor{MainDeltaBetter}{\textuparrow\kern0.15pt 0.19}} & \MainMetricWithDelta{16pt}{15pt}{5.65}{\textcolor{MainDeltaBetter}{\textdownarrow\kern0.15pt 0.05}} & \MainMetricWithDelta{20.5pt}{18pt}{14.96}{\textcolor{MainDeltaWorse}{\textuparrow\kern0.15pt 9.38}} & \MainMetricWithDelta{21.5pt}{18pt}{54.52}{\textcolor{MainDeltaWorse}{\textuparrow\kern0.15pt 51.93}} \\
\hspace*{9pt}GPTQ-W4A16 & \MainMetricWithDelta{20.5pt}{18pt}{53.29}{\textcolor{MainDeltaWorse}{\textdownarrow\kern0.15pt 3.04}} & \MainMetricWithDelta{20.5pt}{18pt}{47.92}{\textcolor{MainDeltaWorse}{\textdownarrow\kern0.15pt 2.22}} & \MainMetricWithDelta{16pt}{15pt}{5.25}{\textcolor{MainDeltaWorse}{\textuparrow\kern0.15pt 0.30}} & \MainMetricWithDelta{20.5pt}{18pt}{60.96}{\textcolor{MainDeltaWorse}{\textdownarrow\kern0.15pt 2.55}} & \MainMetricWithDelta{20.5pt}{18pt}{51.19}{\textcolor{MainDeltaWorse}{\textdownarrow\kern0.15pt 3.13}} & \MainMetricWithDelta{20.5pt}{18pt}{44.19}{\textcolor{MainDeltaWorse}{\textdownarrow\kern0.15pt 1.14}} & \MainMetricWithDelta{16pt}{15pt}{6.16}{\textcolor{MainDeltaWorse}{\textuparrow\kern0.15pt 0.46}} & \MainMetricWithDelta{20.5pt}{18pt}{6.03}{\textcolor{MainDeltaWorse}{\textuparrow\kern0.15pt 0.45}} & \MainMetricWithDelta{21.5pt}{18pt}{9.31}{\textcolor{MainDeltaWorse}{\textuparrow\kern0.15pt 6.72}} \\
\hspace*{9pt}AWQ-W4A16 & \MainMetricWithDelta{20.5pt}{18pt}{56.28}{\textcolor{MainDeltaWorse}{\textdownarrow\kern0.15pt 0.05}} & \MainMetricWithDelta{20.5pt}{18pt}{50.04}{\textcolor{MainDeltaWorse}{\textdownarrow\kern0.15pt 0.10}} & \MainMetricWithDelta{16pt}{15pt}{5.01}{\textcolor{MainDeltaWorse}{\textuparrow\kern0.15pt 0.06}} & \MainMetricWithDelta{20.5pt}{18pt}{63.50}{\textcolor{MainDeltaWorse}{\textdownarrow\kern0.15pt 0.01}} & \MainMetricWithDelta{20.5pt}{18pt}{54.24}{\textcolor{MainDeltaWorse}{\textdownarrow\kern0.15pt 0.08}} & \MainMetricWithDelta{20.5pt}{18pt}{44.97}{\textcolor{MainDeltaWorse}{\textdownarrow\kern0.15pt 0.36}} & \MainMetricWithDelta{16pt}{15pt}{5.48}{\textcolor{MainDeltaBetter}{\textdownarrow\kern0.15pt 0.22}} & \MainMetricWithDelta{20.5pt}{18pt}{6.05}{\textcolor{MainDeltaWorse}{\textuparrow\kern0.15pt 0.47}} & \MainMetricWithDelta{21.5pt}{18pt}{3.21}{\textcolor{MainDeltaWorse}{\textuparrow\kern0.15pt 0.62}} \\
\hspace*{9pt}\textbf{ReCoNav} (IQ3\_XS) & \cellcolor{MainAmber}\MainMetricWithDelta{20.5pt}{18pt}{56.33}{} & \cellcolor{MainAmber}\MainMetricWithDelta{20.5pt}{18pt}{50.14}{} & \cellcolor{MainAmber}\MainMetricWithDelta{16pt}{15pt}{4.95}{} & \cellcolor{MainAmber}\MainMetricWithDelta{20.5pt}{18pt}{63.51}{} & \cellcolor{MainAmber}\MainMetricWithDelta{20.5pt}{18pt}{54.32}{} & \cellcolor{MainAmber}\MainMetricWithDelta{20.5pt}{18pt}{45.33}{} & \cellcolor{MainAmber}\MainMetricWithDelta{16pt}{15pt}{5.70}{} & \cellcolor{MainAmber}\MainMetricWithDelta{20.5pt}{18pt}{5.58}{} & \cellcolor{MainAmber}\MainMetricWithDelta{21.5pt}{18pt}{2.59}{} \\
\rowcolor{MainGray}\multicolumn{10}{l}{\cellcolor{MainGray}\rule{0pt}{2.6ex}\underline{\textit{Uni-NaVid-7B}}} \\
\hspace*{9pt}Full precision & \MainMetricWithDelta{20.5pt}{18pt}{52.15}{\textcolor{MainDeltaBetter}{\textuparrow\kern0.15pt 0.44}} & \MainMetricWithDelta{20.5pt}{18pt}{48.02}{\textcolor{MainDeltaBetter}{\textuparrow\kern0.15pt 0.41}} & \MainMetricWithDelta{16pt}{15pt}{4.96}{\textcolor{MainDeltaBetter}{\textdownarrow\kern0.15pt 0.05}} & \MainMetricWithDelta{20.5pt}{18pt}{57.43}{\textcolor{MainDeltaBetter}{\textuparrow\kern0.15pt 0.12}} & \MainMetricWithDelta{20.5pt}{18pt}{52.32}{\textcolor{MainDeltaBetter}{\textuparrow\kern0.15pt 0.32}} & \MainMetricWithDelta{20.5pt}{18pt}{42.63}{\textcolor{MainDeltaBetter}{\textuparrow\kern0.15pt 0.21}} & \MainMetricWithDelta{16pt}{15pt}{5.67}{\textcolor{MainDeltaBetter}{\textdownarrow\kern0.15pt 0.05}} & \MainMetricWithDelta{20.5pt}{18pt}{14.43}{\textcolor{MainDeltaWorse}{\textuparrow\kern0.15pt 9.33}} & \MainMetricWithDelta{21.5pt}{18pt}{29.97}{\textcolor{MainDeltaWorse}{\textuparrow\kern0.15pt 27.42}} \\
\hspace*{9pt}GPTQ-W4A16 & \MainMetricWithDelta{20.5pt}{18pt}{48.02}{\textcolor{MainDeltaWorse}{\textdownarrow\kern0.15pt 3.69}} & \MainMetricWithDelta{20.5pt}{18pt}{44.27}{\textcolor{MainDeltaWorse}{\textdownarrow\kern0.15pt 3.34}} & \MainMetricWithDelta{16pt}{15pt}{5.57}{\textcolor{MainDeltaWorse}{\textuparrow\kern0.15pt 0.56}} & \MainMetricWithDelta{20.5pt}{18pt}{53.67}{\textcolor{MainDeltaWorse}{\textdownarrow\kern0.15pt 3.64}} & \MainMetricWithDelta{20.5pt}{18pt}{45.95}{\textcolor{MainDeltaWorse}{\textdownarrow\kern0.15pt 6.05}} & \MainMetricWithDelta{20.5pt}{18pt}{38.74}{\textcolor{MainDeltaWorse}{\textdownarrow\kern0.15pt 3.68}} & \MainMetricWithDelta{16pt}{15pt}{7.13}{\textcolor{MainDeltaWorse}{\textuparrow\kern0.15pt 1.41}} & \MainMetricWithDelta{20.5pt}{18pt}{5.57}{\textcolor{MainDeltaWorse}{\textuparrow\kern0.15pt 0.47}} & \MainMetricWithDelta{21.5pt}{18pt}{5.81}{\textcolor{MainDeltaWorse}{\textuparrow\kern0.15pt 3.26}} \\
\hspace*{9pt}AWQ-W4A16 & \MainMetricWithDelta{20.5pt}{18pt}{51.88}{\textcolor{MainDeltaBetter}{\textuparrow\kern0.15pt 0.17}} & \MainMetricWithDelta{20.5pt}{18pt}{47.51}{\textcolor{MainDeltaWorse}{\textdownarrow\kern0.15pt 0.10}} & \MainMetricWithDelta{16pt}{15pt}{4.98}{\textcolor{MainDeltaBetter}{\textdownarrow\kern0.15pt 0.03}} & \MainMetricWithDelta{20.5pt}{18pt}{57.10}{\textcolor{MainDeltaWorse}{\textdownarrow\kern0.15pt 0.21}} & \MainMetricWithDelta{20.5pt}{18pt}{52.04}{\textcolor{MainDeltaBetter}{\textuparrow\kern0.15pt 0.04}} & \MainMetricWithDelta{20.5pt}{18pt}{42.40}{\textcolor{MainDeltaWorse}{\textdownarrow\kern0.15pt 0.02}} & \MainMetricWithDelta{16pt}{15pt}{5.73}{\textcolor{MainDeltaWorse}{\textuparrow\kern0.15pt 0.01}} & \MainMetricWithDelta{20.5pt}{18pt}{5.57}{\textcolor{MainDeltaWorse}{\textuparrow\kern0.15pt 0.47}} & \MainMetricWithDelta{21.5pt}{18pt}{2.83}{\textcolor{MainDeltaWorse}{\textuparrow\kern0.15pt 0.28}} \\
\hspace*{9pt}\textbf{ReCoNav} (IQ3\_XS) & \cellcolor{MainAmber}\MainMetricWithDelta{20.5pt}{18pt}{51.71}{} & \cellcolor{MainAmber}\MainMetricWithDelta{20.5pt}{18pt}{47.61}{} & \cellcolor{MainAmber}\MainMetricWithDelta{16pt}{15pt}{5.01}{} & \cellcolor{MainAmber}\MainMetricWithDelta{20.5pt}{18pt}{57.31}{} & \cellcolor{MainAmber}\MainMetricWithDelta{20.5pt}{18pt}{52.00}{} & \cellcolor{MainAmber}\MainMetricWithDelta{20.5pt}{18pt}{42.42}{} & \cellcolor{MainAmber}\MainMetricWithDelta{16pt}{15pt}{5.72}{} & \cellcolor{MainAmber}\MainMetricWithDelta{20.5pt}{18pt}{5.10}{} & \cellcolor{MainAmber}\MainMetricWithDelta{21.5pt}{18pt}{2.55}{} \\
\rowcolor{MainGray}\multicolumn{10}{l}{\cellcolor{MainGray}\rule{0pt}{2.6ex}\underline{\textit{LightNav-0-4B}}} \\
\hspace*{9pt}Full precision & \MainMetricWithDelta{20.5pt}{18pt}{68.03}{\textcolor{MainDeltaBetter}{\textuparrow\kern0.15pt 0.60}} & \MainMetricWithDelta{20.5pt}{18pt}{63.80}{\textcolor{MainDeltaBetter}{\textuparrow\kern0.15pt 0.04}} & \MainMetricWithDelta{16pt}{15pt}{4.01}{\textcolor{MainDeltaWorse}{\textuparrow\kern0.15pt 0.02}} & \MainMetricWithDelta{20.5pt}{18pt}{70.42}{\textcolor{MainDeltaBetter}{\textuparrow\kern0.15pt 0.11}} & \MainMetricWithDelta{20.5pt}{18pt}{71.14}{\textcolor{MainDeltaBetter}{\textuparrow\kern0.15pt 0.27}} & \MainMetricWithDelta{20.5pt}{18pt}{63.11}{\textcolor{MainDeltaBetter}{\textuparrow\kern0.15pt 0.36}} & \MainMetricWithDelta{16pt}{15pt}{3.87}{\textcolor{MainDeltaBetter}{\textdownarrow\kern0.15pt 0.05}} & \MainMetricWithDelta{20.5pt}{18pt}{8.29}{\textcolor{MainDeltaWorse}{\textuparrow\kern0.15pt 4.74}} & \MainMetricWithDelta{21.5pt}{18pt}{1.53}{\textcolor{MainDeltaWorse}{\textuparrow\kern0.15pt 0.08}} \\
\hspace*{9pt}GPTQ-W4A16 & \MainMetricWithDelta{20.5pt}{18pt}{64.49}{\textcolor{MainDeltaWorse}{\textdownarrow\kern0.15pt 2.94}} & \MainMetricWithDelta{20.5pt}{18pt}{60.35}{\textcolor{MainDeltaWorse}{\textdownarrow\kern0.15pt 3.41}} & \MainMetricWithDelta{16pt}{15pt}{4.20}{\textcolor{MainDeltaWorse}{\textuparrow\kern0.15pt 0.21}} & \MainMetricWithDelta{20.5pt}{18pt}{69.06}{\textcolor{MainDeltaWorse}{\textdownarrow\kern0.15pt 1.25}} & \MainMetricWithDelta{20.5pt}{18pt}{69.04}{\textcolor{MainDeltaWorse}{\textdownarrow\kern0.15pt 1.83}} & \MainMetricWithDelta{20.5pt}{18pt}{61.55}{\textcolor{MainDeltaWorse}{\textdownarrow\kern0.15pt 1.20}} & \MainMetricWithDelta{16pt}{15pt}{4.14}{\textcolor{MainDeltaWorse}{\textuparrow\kern0.15pt 0.22}} & \MainMetricWithDelta{20.5pt}{18pt}{3.32}{\textcolor{MainDeltaBetter}{\textdownarrow\kern0.15pt 0.23}} & \MainMetricWithDelta{21.5pt}{18pt}{5.90}{\textcolor{MainDeltaWorse}{\textuparrow\kern0.15pt 4.45}} \\
\hspace*{9pt}AWQ-W4A16 & \MainMetricWithDelta{20.5pt}{18pt}{67.10}{\textcolor{MainDeltaWorse}{\textdownarrow\kern0.15pt 0.33}} & \MainMetricWithDelta{20.5pt}{18pt}{63.45}{\textcolor{MainDeltaWorse}{\textdownarrow\kern0.15pt 0.31}} & \MainMetricWithDelta{16pt}{15pt}{4.02}{\textcolor{MainDeltaWorse}{\textuparrow\kern0.15pt 0.03}} & \MainMetricWithDelta{20.5pt}{18pt}{70.15}{\textcolor{MainDeltaWorse}{\textdownarrow\kern0.15pt 0.16}} & \MainMetricWithDelta{20.5pt}{18pt}{70.73}{\textcolor{MainDeltaWorse}{\textdownarrow\kern0.15pt 0.14}} & \MainMetricWithDelta{20.5pt}{18pt}{62.66}{\textcolor{MainDeltaWorse}{\textdownarrow\kern0.15pt 0.09}} & \MainMetricWithDelta{16pt}{15pt}{3.96}{\textcolor{MainDeltaWorse}{\textuparrow\kern0.15pt 0.04}} & \MainMetricWithDelta{20.5pt}{18pt}{3.32}{\textcolor{MainDeltaBetter}{\textdownarrow\kern0.15pt 0.23}} & \MainMetricWithDelta{21.5pt}{18pt}{3.20}{\textcolor{MainDeltaWorse}{\textuparrow\kern0.15pt 1.75}} \\
\hspace*{9pt}\textbf{ReCoNav} (Q3\_K\_L) & \cellcolor{MainAmber}\MainMetricWithDelta{20.5pt}{18pt}{67.43}{} & \cellcolor{MainAmber}\MainMetricWithDelta{20.5pt}{18pt}{63.76}{} & \cellcolor{MainAmber}\MainMetricWithDelta{16pt}{15pt}{3.99}{} & \cellcolor{MainAmber}\MainMetricWithDelta{20.5pt}{18pt}{70.31}{} & \cellcolor{MainAmber}\MainMetricWithDelta{20.5pt}{18pt}{70.87}{} & \cellcolor{MainAmber}\MainMetricWithDelta{20.5pt}{18pt}{62.75}{} & \cellcolor{MainAmber}\MainMetricWithDelta{16pt}{15pt}{3.92}{} & \cellcolor{MainAmber}\MainMetricWithDelta{20.5pt}{18pt}{3.55}{} & \cellcolor{MainAmber}\MainMetricWithDelta{21.5pt}{18pt}{1.45}{} \\
\rowcolor{MainGray}\multicolumn{10}{l}{\cellcolor{MainGray}\rule{0pt}{2.6ex}\underline{\textit{SimpleVLN-4B}}} \\
\hspace*{9pt}Full precision & \MainMetricWithDelta{20.5pt}{18pt}{76.27}{\textcolor{MainDeltaBetter}{\textuparrow\kern0.15pt 0.37}} & \MainMetricWithDelta{20.5pt}{18pt}{72.00}{\textcolor{MainDeltaWorse}{\textdownarrow\kern0.15pt 0.04}} & \MainMetricWithDelta{16pt}{15pt}{2.79}{\textcolor{MainDeltaBetter}{\textdownarrow\kern0.15pt 0.01}} & \MainMetricWithDelta{20.5pt}{18pt}{80.04}{\textcolor{MainDeltaBetter}{\textuparrow\kern0.15pt 0.11}} & \MainMetricWithDelta{20.5pt}{18pt}{78.93}{\textcolor{MainDeltaBetter}{\textuparrow\kern0.15pt 0.41}} & \MainMetricWithDelta{20.5pt}{18pt}{69.73}{\textcolor{MainDeltaBetter}{\textuparrow\kern0.15pt 0.37}} & \MainMetricWithDelta{16pt}{15pt}{2.58}{\textcolor{MainDeltaBetter}{\textdownarrow\kern0.15pt 0.02}} & \MainMetricWithDelta{20.5pt}{18pt}{8.46}{\textcolor{MainDeltaWorse}{\textuparrow\kern0.15pt 4.98}} & \MainMetricWithDelta{21.5pt}{18pt}{12.39}{\textcolor{MainDeltaWorse}{\textuparrow\kern0.15pt 0.16}} \\
\hspace*{9pt}GPTQ-W4A16 & \MainMetricWithDelta{20.5pt}{18pt}{75.86}{\textcolor{MainDeltaWorse}{\textdownarrow\kern0.15pt 0.04}} & \MainMetricWithDelta{20.5pt}{18pt}{70.16}{\textcolor{MainDeltaWorse}{\textdownarrow\kern0.15pt 1.88}} & \MainMetricWithDelta{16pt}{15pt}{2.89}{\textcolor{MainDeltaWorse}{\textuparrow\kern0.15pt 0.09}} & \MainMetricWithDelta{20.5pt}{18pt}{78.96}{\textcolor{MainDeltaWorse}{\textdownarrow\kern0.15pt 0.97}} & \MainMetricWithDelta{20.5pt}{18pt}{75.22}{\textcolor{MainDeltaWorse}{\textdownarrow\kern0.15pt 3.30}} & \MainMetricWithDelta{20.5pt}{18pt}{67.00}{\textcolor{MainDeltaWorse}{\textdownarrow\kern0.15pt 2.36}} & \MainMetricWithDelta{16pt}{15pt}{2.92}{\textcolor{MainDeltaWorse}{\textuparrow\kern0.15pt 0.32}} & \MainMetricWithDelta{20.5pt}{18pt}{3.57}{\textcolor{MainDeltaWorse}{\textuparrow\kern0.15pt 0.09}} & \MainMetricWithDelta{21.5pt}{18pt}{20.38}{\textcolor{MainDeltaWorse}{\textuparrow\kern0.15pt 8.15}} \\
\hspace*{9pt}AWQ-W4A16 & \MainMetricWithDelta{20.5pt}{18pt}{75.96}{\textcolor{MainDeltaBetter}{\textuparrow\kern0.15pt 0.06}} & \MainMetricWithDelta{20.5pt}{18pt}{71.89}{\textcolor{MainDeltaWorse}{\textdownarrow\kern0.15pt 0.15}} & \MainMetricWithDelta{16pt}{15pt}{2.83}{\textcolor{MainDeltaWorse}{\textuparrow\kern0.15pt 0.03}} & \MainMetricWithDelta{20.5pt}{18pt}{79.99}{\textcolor{MainDeltaBetter}{\textuparrow\kern0.15pt 0.06}} & \MainMetricWithDelta{20.5pt}{18pt}{78.50}{\textcolor{MainDeltaWorse}{\textdownarrow\kern0.15pt 0.02}} & \MainMetricWithDelta{20.5pt}{18pt}{69.49}{\textcolor{MainDeltaBetter}{\textuparrow\kern0.15pt 0.13}} & \MainMetricWithDelta{16pt}{15pt}{2.62}{\textcolor{MainDeltaWorse}{\textuparrow\kern0.15pt 0.02}} & \MainMetricWithDelta{20.5pt}{18pt}{3.57}{\textcolor{MainDeltaWorse}{\textuparrow\kern0.15pt 0.09}} & \MainMetricWithDelta{21.5pt}{18pt}{14.34}{\textcolor{MainDeltaWorse}{\textuparrow\kern0.15pt 2.11}} \\
\hspace*{9pt}\textbf{ReCoNav} (IQ3\_XS) & \cellcolor{MainAmber}\MainMetricWithDelta{20.5pt}{18pt}{75.90}{} & \cellcolor{MainAmber}\MainMetricWithDelta{20.5pt}{18pt}{72.04}{} & \cellcolor{MainAmber}\MainMetricWithDelta{16pt}{15pt}{2.80}{} & \cellcolor{MainAmber}\MainMetricWithDelta{20.5pt}{18pt}{79.93}{} & \cellcolor{MainAmber}\MainMetricWithDelta{20.5pt}{18pt}{78.52}{} & \cellcolor{MainAmber}\MainMetricWithDelta{20.5pt}{18pt}{69.36}{} & \cellcolor{MainAmber}\MainMetricWithDelta{16pt}{15pt}{2.60}{} & \cellcolor{MainAmber}\MainMetricWithDelta{20.5pt}{18pt}{3.48}{} & \cellcolor{MainAmber}\MainMetricWithDelta{21.5pt}{18pt}{12.23}{} \\
\rowcolor{MainGray}\multicolumn{10}{l}{\cellcolor{MainGray}\rule{0pt}{2.6ex}\underline{\textit{SimpleVLN-9B}}} \\
\hspace*{9pt}Full precision & \MainMetricWithDelta{20.5pt}{18pt}{76.67}{\textcolor{MainDeltaBetter}{\textuparrow\kern0.15pt 0.32}} & \MainMetricWithDelta{20.5pt}{18pt}{72.62}{\textcolor{MainDeltaBetter}{\textuparrow\kern0.15pt 0.12}} & \MainMetricWithDelta{16pt}{15pt}{2.60}{\textcolor{MainDeltaBetter}{\textdownarrow\kern0.15pt 0.09}} & \MainMetricWithDelta{20.5pt}{18pt}{84.07}{\textcolor{MainDeltaBetter}{\textuparrow\kern0.15pt 0.33}} & \MainMetricWithDelta{20.5pt}{18pt}{80.10}{\textcolor{MainDeltaBetter}{\textuparrow\kern0.15pt 0.27}} & \MainMetricWithDelta{20.5pt}{18pt}{69.89}{\textcolor{MainDeltaBetter}{\textuparrow\kern0.15pt 0.42}} & \MainMetricWithDelta{16pt}{15pt}{2.52}{\textcolor{MainDeltaBetter}{\textdownarrow\kern0.15pt 0.06}} & \MainMetricWithDelta{20.5pt}{18pt}{17.53}{\textcolor{MainDeltaWorse}{\textuparrow\kern0.15pt 9.82}} & \MainMetricWithDelta{21.5pt}{18pt}{OOM}{} \\
\hspace*{9pt}GPTQ-W4A16 & \MainMetricWithDelta{20.5pt}{18pt}{76.13}{\textcolor{MainDeltaWorse}{\textdownarrow\kern0.15pt 0.22}} & \MainMetricWithDelta{20.5pt}{18pt}{68.32}{\textcolor{MainDeltaWorse}{\textdownarrow\kern0.15pt 4.18}} & \MainMetricWithDelta{16pt}{15pt}{2.97}{\textcolor{MainDeltaWorse}{\textuparrow\kern0.15pt 0.28}} & \MainMetricWithDelta{20.5pt}{18pt}{80.59}{\textcolor{MainDeltaWorse}{\textdownarrow\kern0.15pt 3.15}} & \MainMetricWithDelta{20.5pt}{18pt}{76.18}{\textcolor{MainDeltaWorse}{\textdownarrow\kern0.15pt 3.65}} & \MainMetricWithDelta{20.5pt}{18pt}{67.53}{\textcolor{MainDeltaWorse}{\textdownarrow\kern0.15pt 1.94}} & \MainMetricWithDelta{16pt}{15pt}{2.83}{\textcolor{MainDeltaWorse}{\textuparrow\kern0.15pt 0.25}} & \MainMetricWithDelta{20.5pt}{18pt}{8.06}{\textcolor{MainDeltaWorse}{\textuparrow\kern0.15pt 0.35}} & \MainMetricWithDelta{21.5pt}{18pt}{34.15}{\textcolor{MainDeltaWorse}{\textuparrow\kern0.15pt 16.99}} \\
\hspace*{9pt}AWQ-W4A16 & \MainMetricWithDelta{20.5pt}{18pt}{76.56}{\textcolor{MainDeltaBetter}{\textuparrow\kern0.15pt 0.21}} & \MainMetricWithDelta{20.5pt}{18pt}{72.54}{\textcolor{MainDeltaBetter}{\textuparrow\kern0.15pt 0.04}} & \MainMetricWithDelta{16pt}{15pt}{2.65}{\textcolor{MainDeltaBetter}{\textdownarrow\kern0.15pt 0.04}} & \MainMetricWithDelta{20.5pt}{18pt}{83.90}{\textcolor{MainDeltaBetter}{\textuparrow\kern0.15pt 0.16}} & \MainMetricWithDelta{20.5pt}{18pt}{79.86}{\textcolor{MainDeltaBetter}{\textuparrow\kern0.15pt 0.03}} & \MainMetricWithDelta{20.5pt}{18pt}{69.53}{\textcolor{MainDeltaBetter}{\textuparrow\kern0.15pt 0.06}} & \MainMetricWithDelta{16pt}{15pt}{2.60}{\textcolor{MainDeltaWorse}{\textuparrow\kern0.15pt 0.02}} & \MainMetricWithDelta{20.5pt}{18pt}{8.06}{\textcolor{MainDeltaWorse}{\textuparrow\kern0.15pt 0.35}} & \MainMetricWithDelta{21.5pt}{18pt}{22.20}{\textcolor{MainDeltaWorse}{\textuparrow\kern0.15pt 5.04}} \\
\hspace*{9pt}\textbf{ReCoNav} (IQ3\_XS) & \cellcolor{MainAmber}\MainMetricWithDelta{20.5pt}{18pt}{76.35}{} & \cellcolor{MainAmber}\MainMetricWithDelta{20.5pt}{18pt}{72.50}{} & \cellcolor{MainAmber}\MainMetricWithDelta{16pt}{15pt}{2.69}{} & \cellcolor{MainAmber}\MainMetricWithDelta{20.5pt}{18pt}{83.74}{} & \cellcolor{MainAmber}\MainMetricWithDelta{20.5pt}{18pt}{79.83}{} & \cellcolor{MainAmber}\MainMetricWithDelta{20.5pt}{18pt}{69.47}{} & \cellcolor{MainAmber}\MainMetricWithDelta{16pt}{15pt}{2.58}{} & \cellcolor{MainAmber}\MainMetricWithDelta{20.5pt}{18pt}{7.71}{} & \cellcolor{MainAmber}\MainMetricWithDelta{21.5pt}{18pt}{17.16}{} \\
\bottomrule
\end{tabularx}

\endgroup
\end{table*}
\endgroup

\begin{figure*}[!t]
  \centering
  \includegraphics[width=\textwidth]{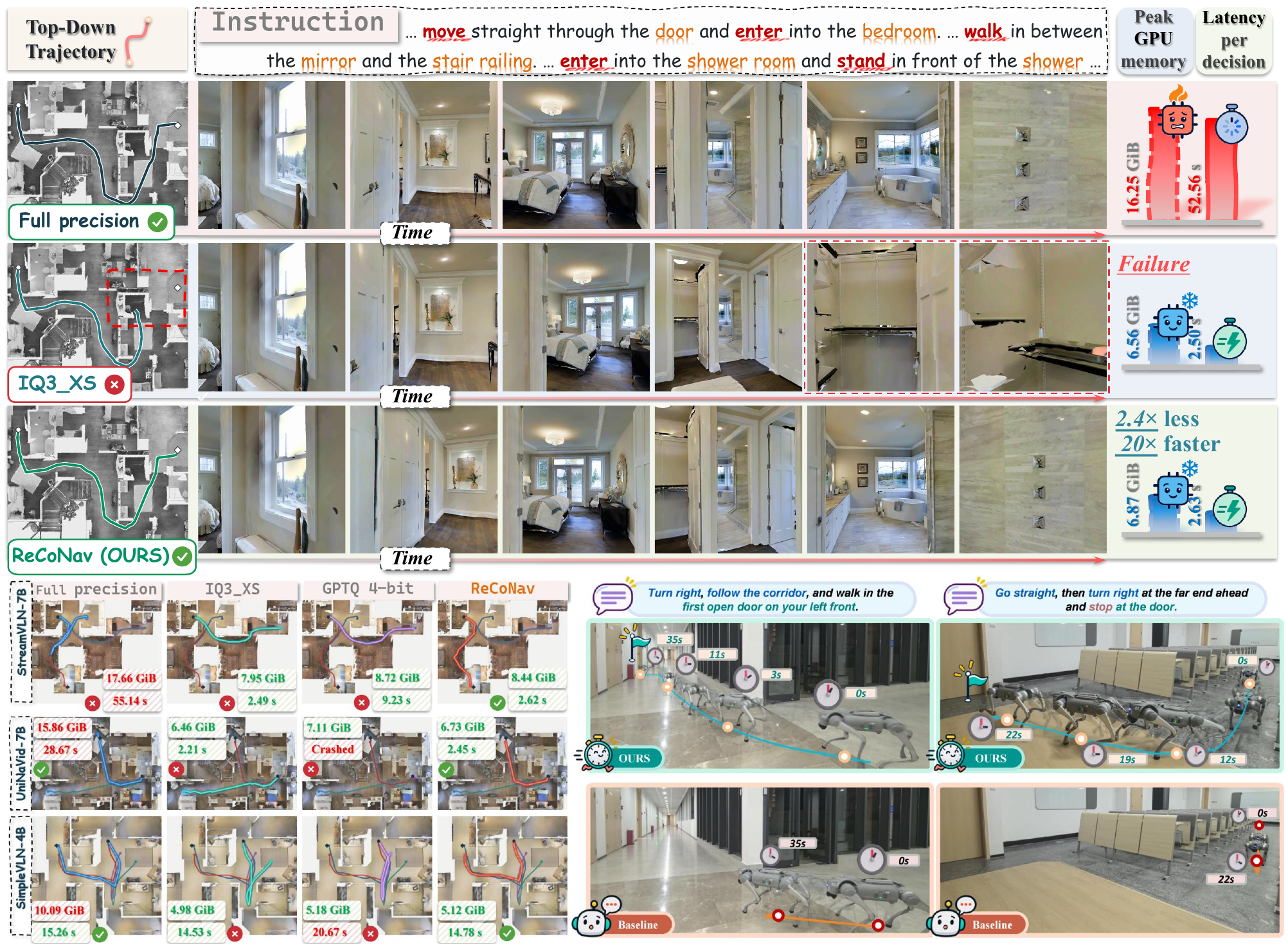}
\caption{\textbf{Closed-loop navigation in simulation and on a Unitree Go2.} Top and bottom left: selected simulation episodes, with peak GPU memory measured on NVIDIA A800 and mean per-decision latency on Jetson Orin NX 16GB. Bottom right: robot trajectories and elapsed task time for ReCoNav and the full-precision deployment. Checkmarks and crosses indicate task success and failure.}
  \label{fig:showcase_sim}
\end{figure*}

\subsection{Experimental Setup}
\label{sec:experimental-setup}

\paragraph{Navigation evaluation.}
We evaluate closed-loop navigation on the unseen validation splits of R2R and RxR in VLN-CE~\cite{anderson2018vision,krantz2020beyond,ku2020room}, using English instructions. StreamVLN-7B~\cite{wei2025streamvln} and Uni-NaVid-7B~\cite{zhang2025uninavid} are our primary backbones; LightNav-0-4B~\cite{wang2026lightnav} and SimpleVLN-4B/9B~\cite{hong2026simplevln} extend evaluation to spatial-pointer and text-command interfaces. For each configuration, we compare full precision, GPTQ-W4A16~\cite{frantar2023gptq}, AWQ-W4A16~\cite{lin2024awq}, and ReCoNav on the unseen splits. Both 4-bit baselines use a group size of 128. ReCoNav uses \texttt{Q3\_K\_L} for LightNav-0 and \texttt{IQ3\_XS} otherwise. We report success rate (SR), success weighted by path length (SPL), and navigation error (NE) on both datasets, and oracle success (OS) on R2R.

\paragraph{On-device inference.}
We evaluate navigation quality through closed-loop Habitat~\cite{savva2019habitat} rollouts on an NVIDIA A800 server and replay each deployment's recorded decision inputs on an NVIDIA Jetson Orin NX 16GB mounted on a Unitree Go2. Full-precision models use PyTorch, whereas ReCoNav integrates packed low-bit inference and residual compensation into llama.cpp~\cite{gerganov2024llamacpp}. GPTQ and AWQ use model-specific adaptations of their respective implementations. Decision latency spans RGB availability to action output, including triggered reruns. Peak GPU memory is measured separately on an NVIDIA A800. Sec.~\ref{app:setup} gives the timing protocols.

\begin{figure*}[!t]
\centering
\includegraphics[trim=0 9bp 0 9bp,clip,width=0.82\textwidth]{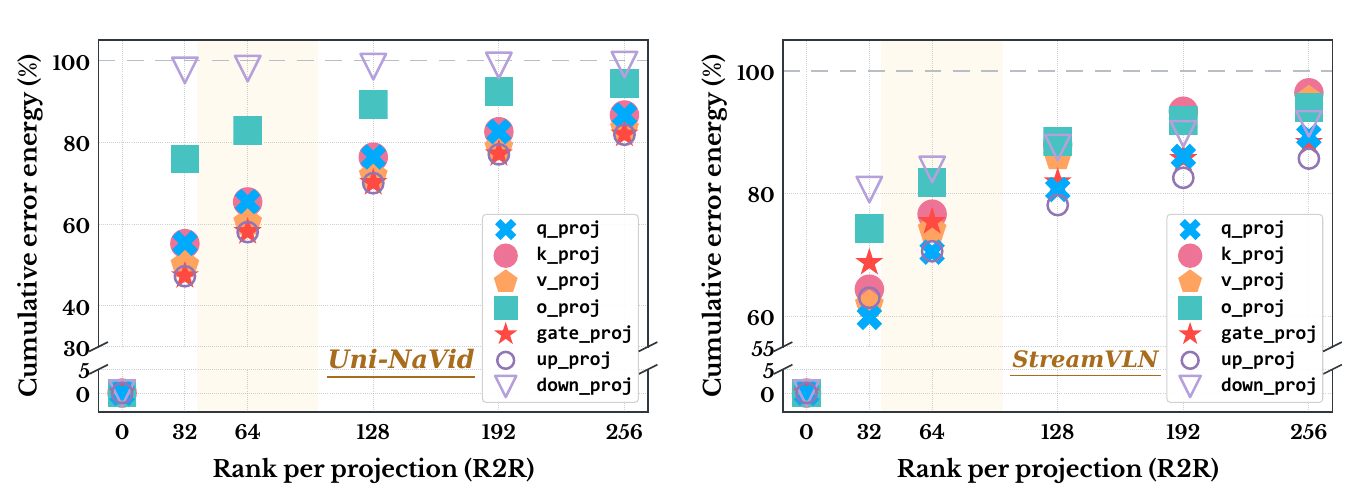}
\caption{Cumulative residual energy on R2R for Uni-NaVid (left) and StreamVLN (right), pooled by projection type across layers. Different energy capture at equal ranks motivates mixed-rank allocation.}
\label{fig:residual-spectrum}
\end{figure*}

\begin{table*}[t]
\centering
\captionsetup{skip=5pt}
\caption{\textbf{Effect of compensation scheduling on navigation performance.} Comparisons on R2R and RxR unseen splits use fixed IQ3\_XS weights and correction factors within each backbone--dataset pair.}
\label{tab:compensation-navigation-v5}
\begingroup
\small
\setlength{\tabcolsep}{5pt}
\setlength{\aboverulesep}{0pt}
\setlength{\belowrulesep}{0pt}
\renewcommand{\arraystretch}{1.00}
\begin{tabularx}{\linewidth}{p{50pt}p{91pt}*{7}{N}}
\toprule
\multirow{2}{*}[-0.634pt]{\textbf{Backbone}} & \multirow{2}{*}[-0.634pt]{\textbf{Compensation}} & \multicolumn{4}{c}{\textbf{R2R unseen}} & \multicolumn{3}{c}{\textbf{RxR unseen}} \\
\cmidrule(l{\tabcolsep}r{\tabcolsep}){3-6}\cmidrule(l{\tabcolsep}r{\tabcolsep}){7-9}
 &  & \multicolumn{1}{>{\columncolor{NavMetricHeader}[0pt][\tabcolsep]}r}{SR $\uparrow$} & \multicolumn{1}{>{\columncolor{NavMetricHeader}[\tabcolsep][\tabcolsep]}r}{SPL $\uparrow$} & \multicolumn{1}{>{\columncolor{NavMetricHeader}[\tabcolsep][\tabcolsep]}r}{NE $\downarrow$} & \multicolumn{1}{>{\columncolor{NavMetricHeader}[\tabcolsep][0pt]}r}{OS $\uparrow$} & \multicolumn{1}{>{\columncolor{NavMetricHeader}[0pt][\tabcolsep]}r}{SR $\uparrow$} & \multicolumn{1}{>{\columncolor{NavMetricHeader}[\tabcolsep][\tabcolsep]}r}{SPL $\uparrow$} & \multicolumn{1}{>{\columncolor{NavMetricHeader}[\tabcolsep][0pt]}r}{NE $\downarrow$} \\
\midrule
\multirow{4}{*}[-0.634pt]{\textbf{Uni-NaVid}} & No compensation & 46.38 & 43.20 & 5.60 & 51.01 & 46.91 & 38.30 & 7.35 \\
 & Every decision & \textbf{51.71} & \textbf{47.68} & \textbf{4.98} & \textbf{57.37} & \textbf{52.09} & \textbf{42.49} & \textbf{5.69} \\
 & Random selection & 50.03 & 46.71 & 5.32 & 56.17 & 51.13 & 41.33 & 6.02 \\
\cmidrule(l{\tabcolsep}r{\tabcolsep}){2-9}
 & \multicolumn{1}{>{\columncolor{NavCoral}[0pt][\tabcolsep]}l}{\NavOursName} & \cellcolor{NavCoral}\textbf{51.71} & \cellcolor{NavCoral}\underline{47.61} & \cellcolor{NavCoral}\underline{5.01} & \cellcolor{NavCoral}\underline{57.31} & \cellcolor{NavCoral}\underline{52.00} & \cellcolor{NavCoral}\underline{42.42} & \multicolumn{1}{>{\columncolor{NavCoral}[\tabcolsep][0pt]}r}{\underline{5.72}} \\
\midrule
\multirow{4}{*}[-0.634pt]{\textbf{StreamVLN}} & No compensation & 54.00 & 47.80 & 5.09 & 62.21 & 50.20 & 43.00 & 6.39 \\
 & Every decision & \textbf{56.39} & \textbf{50.20} & \textbf{4.92} & \textbf{63.62} & \textbf{54.40} & \textbf{45.40} & \textbf{5.67} \\
 & Random selection & 55.25 & 49.92 & 5.02 & 62.81 & 53.42 & 44.76 & 5.89 \\
\cmidrule(l{\tabcolsep}r{\tabcolsep}){2-9}
 & \multicolumn{1}{>{\columncolor{NavCoral}[0pt][\tabcolsep]}l}{\NavOursName} & \cellcolor{NavCoral}\underline{56.33} & \cellcolor{NavCoral}\underline{50.14} & \cellcolor{NavCoral}\underline{4.95} & \cellcolor{NavCoral}\underline{63.51} & \cellcolor{NavCoral}\underline{54.32} & \cellcolor{NavCoral}\underline{45.33} & \multicolumn{1}{>{\columncolor{NavCoral}[\tabcolsep][0pt]}r}{\underline{5.70}} \\
\bottomrule
\end{tabularx}
\endgroup
\end{table*}

\subsection{Navigation Quality and Deployment}
\label{sec:main-results}

\paragraph{Main navigation results.}
Across the two primary backbones and both datasets, ReCoNav retains success rates within 0.44 percentage points of full precision while reducing model storage by over 60\% (Tab.~\ref{tab:main}). The complete low-bit deployments achieve $11.8$--$21.1\times$ speedups in on-device decision inference over full precision. Across all five configurations, ReCoNav improves every reported navigation metric over GPTQ-W4A16 while retaining SR comparable to AWQ-W4A16. ReCoNav also achieves lower mean decision latency than both four-bit baselines, with reductions of $9.9$--$19.3\%$ relative to AWQ on the two primary backbones.

In Fig.~\ref{fig:showcase_sim} (top), ReCoNav reaches the goal where uncompensated \texttt{IQ3\_XS} fails, with lower peak GPU memory and decision latency than full precision. Further examples in the bottom-left panel show ReCoNav successfully navigating episodes in which uncompensated quantization and GPTQ-W4A16 fail. For Uni-NaVid, \emph{Crashed} indicates that GPTQ-W4A16 fails to produce a valid action in time. Sec.~\ref{app:qualitative-discussion} extends the trajectory analysis to all five configurations.

\paragraph{Transfer across action interfaces.}
ReCoNav extends to spatial-pointer and text-command models, keeping SR within 0.60 percentage points of full precision on LightNav-0 and SimpleVLN-4B/9B across both datasets (Tab.~\ref{tab:main}). Both 4B configurations more than halve model storage while retaining latency close to full precision. Although LightNav-0 uses 0.23 GiB more storage than either four-bit baseline, it achieves a mean decision latency of $1.45$~s, compared with $3.20$~s for AWQ and $5.90$~s for GPTQ.

\paragraph{Larger-model deployment.}
ReCoNav fits SimpleVLN-9B within the memory budget of Jetson Orin NX 16GB, using 7.71 GiB of model storage and 17.16 s per decision (Tab.~\ref{tab:main}), compared with $22.20$~s for AWQ and $34.15$~s for GPTQ. Full precision exceeds device memory; its navigation metrics come from server rollouts.

\subsection{Compensation Scheduling}
\label{sec:allocation}

\begin{figure*}[!t]
\centering
\includegraphics[trim=0 9bp 0 13bp,clip,width=\textwidth]{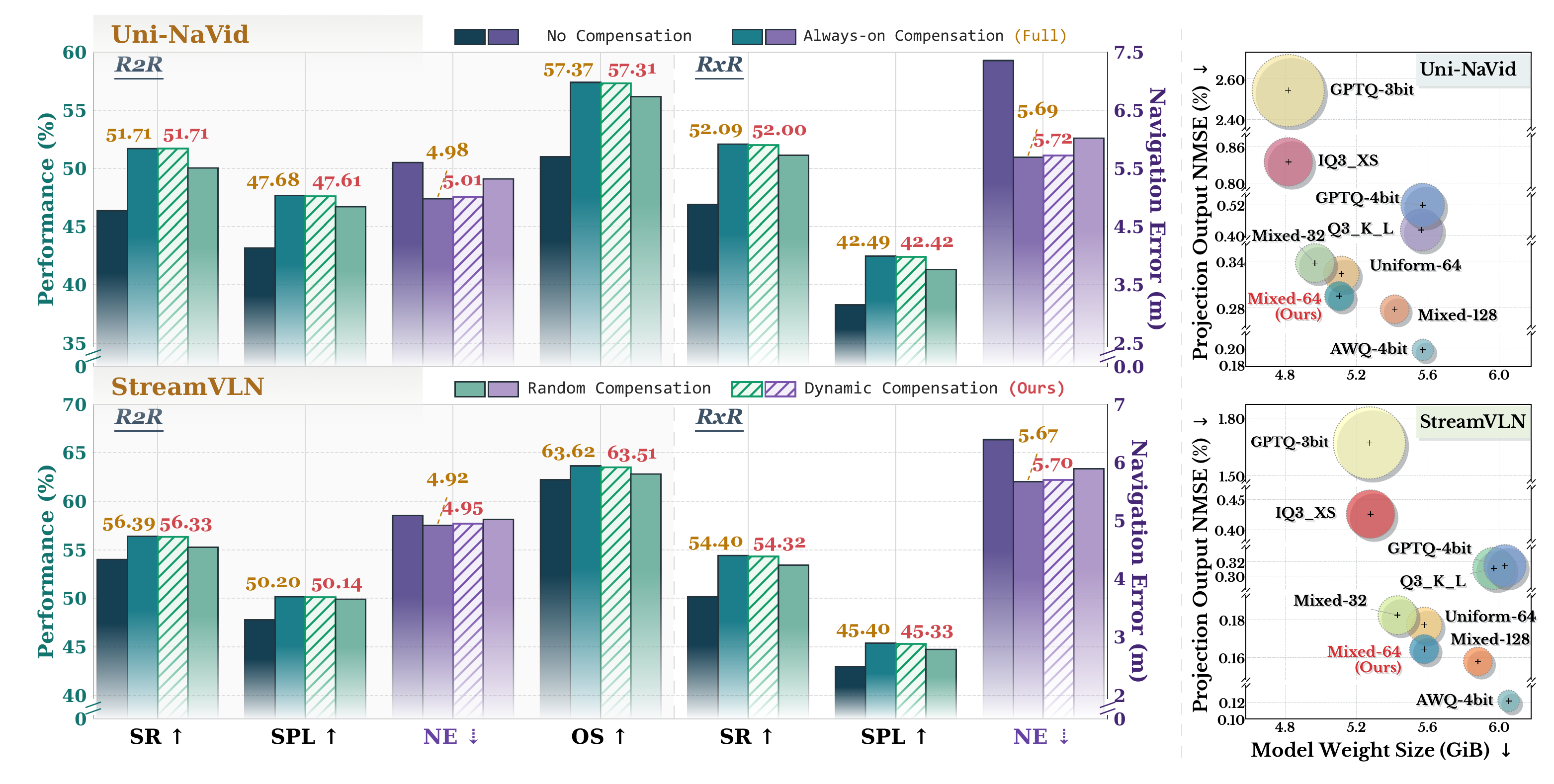}
\caption{Scheduling and rank ablations for Uni-NaVid (top) and StreamVLN (bottom). Left: R2R/RxR navigation performance with fixed quantized weights and corrections. Right: projection-output NMSE versus model weight size across quantization and rank settings. Lower-left positions indicate lower reconstruction error and smaller model weight size.}
\label{fig:rank-memory}
\end{figure*}

With \texttt{IQ3\_XS} weights and correction factors fixed, ReCoNav improves SR by 2.33--5.33 percentage points over uncompensated inference and remains within 0.09 points of always-on compensation across both backbones and datasets (Tab.~\ref{tab:compensation-navigation-v5} and Fig.~\ref{fig:rank-memory}, left panels). On mixed R2R/RxR StreamVLN episodes, ReCoNav compensates only 39.8\% of decisions, reducing mean decision latency from 2.89 to 2.59 s (10.4\%) relative to always-on compensation (Tab.~\ref{tab:scheduling}).

\begin{table}[H]
\centering
\caption{\textbf{Compensation scheduling on Jetson Orin NX.} StreamVLN is evaluated on mixed R2R/RxR episodes. Comp. and Rerun denote the fractions of decisions accepted with compensation and requiring recomputation, respectively. Latency includes reruns.}
\label{tab:scheduling}
\begingroup\small
\setlength{\tabcolsep}{2.5pt}
\setlength{\aboverulesep}{0pt}\setlength{\belowrulesep}{0pt}
\renewcommand{\arraystretch}{1.22}
\renewcommand{\tabularxcolumn}[1]{m{#1}}
\begin{tabularx}{\linewidth}{>{\raggedright\arraybackslash}Xrrr}
\toprule
\textbf{Policy} & Comp. (\%) & Rerun (\%) $\downarrow$ & Latency (s/dec.) $\downarrow$ \\
\midrule
No compensation & 0.0 & 0.0 & 2.48 \\
Every decision & 100.0 & 0.0 & 2.89 \\
Random selection & 39.8 & 23.0 & 2.67 \\
\rowcolor{NavCoral}\textbf{Dynamic (ours)} & 39.8 & 15.9 & 2.59 \\
\bottomrule
\end{tabularx}

\endgroup
\end{table}

\paragraph{Compensation timing.}
To assess compensation timing, we replace only the uncertainty tests in ReCoNav with random draws. For each comparison, we set $p$ to the dynamic policy's measured compensation rate and use independent Bernoulli($p$) draws for both rerun and continuation decisions, without further tuning. A positive draw triggers a rerun after uncompensated generation or retains compensation after compensated generation. ReCoNav improves every reported navigation metric over this control, with SR gains of 0.87--1.68 percentage points (Tab.~\ref{tab:compensation-navigation-v5}).

Both policies record a compensation rate of 39.8\% on StreamVLN at the reported precision. ReCoNav reduces the rerun rate from 23.0\% to 15.9\% and mean latency from 2.67 to 2.59 s (Tab.~\ref{tab:scheduling}). About 60\% of ReCoNav's compensated decisions start directly with compensation enabled, avoiding an initial uncompensated generation. For the remaining reruns, draft verification accelerates compensated language-model recomputation by 1.49--3.29$\times$ across the five model configurations on A800 (Sec.~\ref{app:draft-evaluation}).

\subsection{Fixed-Budget Rank Allocation}
\label{sec:rank-memory}

Equal ranks capture different fractions of residual energy across projection types in both backbones (\cref{fig:residual-spectrum}). For example, Uni-NaVid's feed-forward down projections capture most residual energy at low rank, whereas several other projections continue to benefit from higher ranks. This variation motivates allocating capacity by reconstruction benefit under a shared storage budget.

Mixed-$k$ uses the factor storage of Uniform-$k$ as its budget ceiling. We compare Mixed-64 and Uniform-64 using identical fitting data and held-out training trajectories. Mixed-64 follows the budgeted allocation in Sec.~\ref{sec:vlnq-reconstruction}, accounting for projection dimensions in addition to the mean-rank cap. We measure projection-output NMSE as the summed squared reconstruction residual normalized by the summed squared full-precision outputs. Mixed-64 lowers NMSE on both backbones without increasing model size (Fig.~\ref{fig:rank-memory}, right panels).

\subsection{Robot Deployment}
\label{sec:robot-deployment}

On a Unitree Go2, StreamVLN with ReCoNav completes two closed-loop navigation tasks in 35 s and 22 s, respectively (Fig.~\ref{fig:showcase_sim}, bottom right). The full-precision deployment makes limited progress over the corresponding time intervals. These examples illustrate how shorter decision delays support sustained navigation through task completion.

\section{Conclusion}
\label{sec:conclusion}

ReCoNav uses action uncertainty to schedule low-rank compensation on a fixed quantized navigation backbone, recomputing uncertain action candidates before execution and carrying compensation into subsequent decisions when needed to avoid redundant uncompensated passes. During offline preparation, ReCoNav fits activation-weighted corrections and allocates their ranks across projections to reduce reconstruction error under a fixed storage budget. Across five model configurations on R2R and RxR, ReCoNav maintains near-full-precision navigation success while reducing model storage by over 60\% on StreamVLN and Uni-NaVid. The scheduling results show that the navigation benefits of always-on compensation can be largely retained by selectively compensating a subset of decisions. On Jetson Orin NX, the complete quantized StreamVLN deployment achieves approximately $20\times$ faster decision inference than full precision, while quantized SimpleVLN-9B fits within the device memory budget.

\label{main-text-end}

\clearpage

\section*{LLM Usage Statement}
We used ChatGPT to assist with language polishing, refining related work and citations, code editing, and figure refinement. All research ideas, methods, and experimental results are the authors’ original work. The authors have reviewed the manuscript and take full responsibility for its content and conclusions.

{
  \small
  \bibliographystyle{ieeenat_fullname}
  \bibliography{main}

\begin{thebibliography}{24}
\providecommand{\natexlab}[1]{#1}
\providecommand{\url}[1]{\texttt{#1}}
\expandafter\ifx\csname urlstyle\endcsname\relax
  \providecommand{\doi}[1]{doi: #1}\else
  \providecommand{\doi}{doi: \begingroup \urlstyle{rm}\Url}\fi

\bibitem[Anderson et~al.(2018)Anderson, Wu, Teney, Bruce, Johnson, S{\"u}nderhauf, Reid, Gould, and van~den Hengel]{anderson2018vision}
Peter Anderson, Qi Wu, Damien Teney, Jake Bruce, Mark Johnson, Niko S{\"u}nderhauf, Ian Reid, Stephen Gould, and Anton van~den Hengel.
\newblock Vision-and-language navigation: Interpreting visually-grounded navigation instructions in real environments.
\newblock In \emph{IEEE/CVF Conference on Computer Vision and Pattern Recognition (CVPR)}, pages 3674--3683. IEEE, 2018.

\bibitem[Chen et~al.(2021)Chen, Guhur, Schmid, and Laptev]{chen2021hamt}
Shizhe Chen, Pierre-Louis Guhur, Cordelia Schmid, and Ivan Laptev.
\newblock History aware multimodal transformer for vision-and-language navigation.
\newblock In \emph{NeurIPS}, pages 5834--5847, 2021.

\bibitem[Chen et~al.(2022)Chen, Guhur, Tapaswi, Schmid, and Laptev]{chen2022duet}
Shizhe Chen, Pierre-Louis Guhur, Makarand Tapaswi, Cordelia Schmid, and Ivan Laptev.
\newblock Think global, act local: Dual-scale graph transformer for vision-and-language navigation.
\newblock In \emph{CVPR}, pages 16516--16526, 2022.

\bibitem[Frantar et~al.(2023)Frantar, Ashkboos, Hoefler, and Alistarh]{frantar2023gptq}
Elias Frantar, Saleh Ashkboos, Torsten Hoefler, and Dan Alistarh.
\newblock {OPTQ}: Accurate quantization for generative pre-trained transformers.
\newblock In \emph{The Eleventh International Conference on Learning Representations}, 2023.

\bibitem[{Georgi Gerganov and the llama.cpp contributors}(2026)]{gerganov2024llamacpp}
{Georgi Gerganov and the llama.cpp contributors}.
\newblock llama.cpp: {LLM} inference in {C/C++}.
\newblock \url{https://github.com/ggml-org/llama.cpp}, 2026.
\newblock Revision 434ddbbc, September 9, 2026.

\bibitem[Hong et~al.(2026)Hong, Cai, Chen, Zhou, He, Guo, Sun, Ma, Yang, and Zhang]{hong2026simplevln}
Lingyi Hong, Weiren Cai, Jiayu Chen, Xinyu Zhou, Bailin He, Yunfei Guo, Qiao Sun, Boyu Ma, Jianfei Yang, and Wenqiang Zhang.
\newblock Simplevln: Towards simple and frontier vision-language navigation, 2026.

\bibitem[Krantz et~al.(2020)Krantz, Wijmans, Majumdar, Batra, and Lee]{krantz2020beyond}
Jacob Krantz, Erik Wijmans, Arjun Majumdar, Dhruv Batra, and Stefan Lee.
\newblock Beyond the nav-graph: Vision-and-language navigation in continuous environments.
\newblock In \emph{ECCV}, pages 104--120. Springer, 2020.

\bibitem[Ku et~al.(2020)Ku, Anderson, Patel, Ie, and Baldridge]{ku2020room}
Alexander Ku, Peter Anderson, Roma Patel, Eugene Ie, and Jason Baldridge.
\newblock Room-across-room: Multilingual vision-and-language navigation with dense spatiotemporal grounding.
\newblock In \emph{Proceedings of the 2020 Conference on Empirical Methods in Natural Language Processing (EMNLP)}, pages 4392--4412, 2020.

\bibitem[Lin et~al.(2024)Lin, Tang, Tang, Yang, Chen, Wang, Xiao, Dang, Gan, and Han]{lin2024awq}
Ji Lin, Jiaming Tang, Haotian Tang, Shang Yang, Wei-Ming Chen, Wei-Chen Wang, Guangxuan Xiao, Xingyu Dang, Chuang Gan, and Song Han.
\newblock {AWQ}: Activation-aware weight quantization for on-device {LLM} compression and acceleration.
\newblock In \emph{Proceedings of Machine Learning and Systems}, pages 87--100, 2024.

\bibitem[Liu et~al.(2026)Liu, Li, Que, Yang, Shao, and Hu]{liu2026spear}
Hongyuan Liu, Yawei Li, Zhiqiang Que, Qinli Yang, Junming Shao, and Guosheng Hu.
\newblock {SPEAR}: A system for post-quantization error-adaptive recovery enabling efficient low-bit {LLM} serving.
\newblock arXiv:2606.11244, 2026.

\bibitem[Park et~al.(2025)Park, Hyun, Kim, and Lee]{park2025decdec}
Yeonhong Park, Jake Hyun, Hojoon Kim, and Jae~W. Lee.
\newblock {DecDEC}: A systems approach to advancing low-bit {LLM} quantization.
\newblock In \emph{OSDI}, pages 803--819, 2025.

\bibitem[Savva et~al.(2019)Savva, Kadian, Maksymets, Zhao, Wijmans, Jain, Straub, Liu, Koltun, Malik, Parikh, and Batra]{savva2019habitat}
Manolis Savva, Abhishek Kadian, Oleksandr Maksymets, Yili Zhao, Erik Wijmans, Bhavana Jain, Julian Straub, Jia Liu, Vladlen Koltun, Jitendra Malik, Devi Parikh, and Dhruv Batra.
\newblock {Habitat}: A platform for embodied {AI} research.
\newblock In \emph{Proceedings of the IEEE/CVF International Conference on Computer Vision (ICCV)}, pages 9339--9347, 2019.

\bibitem[Shao et~al.(2024)Shao, Chen, Zhang, Xu, Zhao, Li, Zhang, Peng, Qiao, and Luo]{shao2024omniquant}
Wenqi Shao, Mengzhao Chen, Zhaoyang Zhang, Peng Xu, Lirui Zhao, Zhiqian Li, Kaipeng Zhang, Gao Peng, Yu Qiao, and Ping Luo.
\newblock {OmniQuant}: Omnidirectionally calibrated quantization for large language models.
\newblock In \emph{International Conference on Learning Representations}, pages 45472--45496, 2024.

\bibitem[Wang et~al.(2026)Wang, Luo, Huang, Xu, Wang, Wang, Ma, Yang, Mei, Wei, Hu, Liu, Chen, Shao, Lin, Li, Pan, Liu, Ma, and Fan]{wang2026lightnav}
Shaoan Wang, Aocheng Luo, Fei Huang, Jingyi Xu, Xiaoyang Wang, Yueyu Wang, Qianli Ma, Fan Yang, Ran Mei, Jia Wei, Jiangpeng Hu, Xuhao Liu, Hongming Chen, Yuanbin Shao, Yiyang Lin, Ziliang Li, Liang Pan, Xinhang Liu, Yuntao Ma, and Tingxiang Fan.
\newblock Lightnav-0: Eliciting vlm spatial intelligence for generalist embodied navigation, 2026.

\bibitem[Wei et~al.(2026)Wei, Wan, Yu, Wang, Mao, Zhu, Cai, Wang, Chen, Liu, and Pang]{wei2025streamvln}
Meng Wei, Chenyang Wan, Xiqian Yu, Tai Wang, Xiaohan Mao, Chenming Zhu, Wenzhe Cai, Hanqing Wang, Yilun Chen, Xihui Liu, and Jiangmiao Pang.
\newblock {StreamVLN}: Streaming vision-and-language navigation via {SlowFast} context modeling.
\newblock In \emph{2026 IEEE International Conference on Robotics and Automation (ICRA)}, pages 22174--22181. IEEE, 2026.

\bibitem[Xiao et~al.(2023)Xiao, Lin, Seznec, Wu, Demouth, and Han]{xiao2023smoothquant}
Guangxuan Xiao, Ji Lin, Mickael Seznec, Hao Wu, Julien Demouth, and Song Han.
\newblock {SmoothQuant}: Accurate and efficient post-training quantization for large language models.
\newblock In \emph{ICML}, pages 38087--38099, 2023.

\bibitem[Xu et~al.(2026)Xu, Yang, Fan, Liu, Li, Li, and Zhang]{xu2026qvla}
Yuhao Xu, Yantai Yang, Zhenyang Fan, Yufan Liu, Yuming Li, Bing Li, and Zhipeng Zhang.
\newblock {QVLA}: Not all channels are equal in vision-language-action model\textquotesingle s quantization.
\newblock In \emph{International Conference on Learning Representations}, pages 154256--154268, 2026.

\bibitem[Yao et~al.(2022)Yao, Yazdani~Aminabadi, Zhang, Wu, Li, and He]{yao2022zeroquant}
Zhewei Yao, Reza Yazdani~Aminabadi, Minjia Zhang, Xiaoxia Wu, Conglong Li, and Yuxiong He.
\newblock {ZeroQuant}: Efficient and affordable post-training quantization for large-scale transformers.
\newblock In \emph{NeurIPS}, pages 27168--27183, 2022.

\bibitem[Zhang et~al.(2024{\natexlab{a}})Zhang, Cheng, Constantinides, and Zhao]{zhang2024lqer}
Cheng Zhang, Jianyi Cheng, George~A. Constantinides, and Yiren Zhao.
\newblock {LQER}: Low-rank quantization error reconstruction for {LLMs}.
\newblock In \emph{ICML}, pages 58763--58779, 2024{\natexlab{a}}.

\bibitem[Zhang et~al.(2025{\natexlab{a}})Zhang, Wong, Xiao, Constantinides, and Zhao]{zhang2025qera}
Cheng Zhang, Jeffrey T.~H. Wong, Can Xiao, George Constantinides, and Yiren Zhao.
\newblock {QERA}: an analytical framework for quantization error reconstruction.
\newblock In \emph{International Conference on Learning Representations}, pages 12531--12560, 2025{\natexlab{a}}.

\bibitem[Zhang et~al.(2024{\natexlab{b}})Zhang, Wang, Xu, Zhou, Hong, Fang, Wu, Zhang, and Wang]{zhang2024navid}
Jiazhao Zhang, Kunyu Wang, Rongtao Xu, Gengze Zhou, Yicong Hong, Xiaomeng Fang, Qi Wu, Zhizheng Zhang, and He Wang.
\newblock {NaVid}: Video-based {VLM} plans the next step for vision-and-language navigation.
\newblock In \emph{RSS}, 2024{\natexlab{b}}.

\bibitem[Zhang et~al.(2025{\natexlab{b}})Zhang, Wang, Wang, Li, Liu, Wei, Wang, Zhang, and Wang]{zhang2025uninavid}
Jiazhao Zhang, Kunyu Wang, Shaoan Wang, Minghan Li, Haoran Liu, Songlin Wei, Zhongyuan Wang, Zhizheng Zhang, and He Wang.
\newblock {Uni-NaVid}: A video-based vision-language-action model for unifying embodied navigation tasks.
\newblock In \emph{RSS}, 2025{\natexlab{b}}.

\bibitem[Zhang et~al.(2026)Zhang, Hsieh, Wan, Lin, Wang, Wang, Lei, and Zhang]{zhang2026quantvla}
Jingxuan Zhang, Yunta Hsieh, Zhongwei Wan, Haokun Lin, Xin Wang, Ziqi Wang, Yingtie Lei, and Mi Zhang.
\newblock {QuantVLA}: Scale-calibrated post-training quantization for vision-language-action models.
\newblock In \emph{CVPR}, pages 39539--39549, 2026.

\bibitem[Zheng et~al.(2026)Zheng, Cao, Tao, Mao, Zhang, Li, Chen, Zou, Liu, Cao, Mei, and Chen]{zheng2026dyqvla}
Zihao Zheng, Hangyu Cao, Chibang Tao, Zhihao Mao, Lingyue Zhang, Maoliang Li, Jiayu Chen, Hailong Zou, Xuanzhe Liu, Donggang Cao, Hong Mei, and Xiang Chen.
\newblock {DyQ-VLA}: Temporal-dynamic-aware quantization for embodied vision-language-action models.
\newblock arXiv:2603.07904v3, 2026.

\end{thebibliography}
}
\clearpage
\makeatletter
\setlength{\@fptop}{0pt}
\setlength{\@dblfptop}{0pt}
\makeatother
\definecolor{IOBrown}{HTML}{644734}
\definecolor{IOTitle}{HTML}{F2E6DA}
\definecolor{IOEdge}{HTML}{DCCAAB}
\definecolor{IOCard}{HTML}{FFFCF9}
\definecolor{IOAmber}{HTML}{F8E5BE}
\definecolor{IOScorePaper}{HTML}{FFF7E9}
\definecolor{IOCoral}{HTML}{F9E5DE}
\definecolor{IOCoralInk}{HTML}{853F32}
\definecolor{IOAgree}{HTML}{F3ECE3}
\definecolor{IOMuted}{HTML}{736B64}
\newcommand{\ioSmallCaps}[1]{{\usefont{T1}{ptm}{m}{sc}#1}}
\newtcolorbox{nativecard}[2]{%
  enhanced,sharp corners,colback=IOCard,colframe=IOEdge,colbacktitle=IOTitle,
  coltitle=IOBrown,boxrule=0.35pt,boxsep=0pt,left=7pt,right=7pt,
  top=5pt,bottom=5pt,toptitle=3.7pt,bottomtitle=3.7pt,
  borderline west={0.9pt}{0pt}{IOBrown!70},
  title={\ioSmallCaps{#1}\hfill{\footnotesize\itshape #2}},
  fonttitle=\fontsize{10}{11.2}\selectfont,
  fontupper=\fontsize{8.2}{9.4}\selectfont}
\newcommand{\iotoken}[1]{\texttt{\detokenize{#1}}}
\newcommand{\ioIgnored}[1]{\textcolor{IOMuted}{#1}}
\newcommand{\ioScored}[1]{{\setlength{\fboxsep}{1pt}%
  \colorbox{IOAmber}{\textcolor{IOBrown}{#1}}}}
\newcommand{\ioChanged}[1]{\textcolor{IOCoralInk}{\textbf{#1}}}
\newcommand{\ioCode}[1]{{\fontsize{7.7}{8.7}\selectfont #1}}
\newcommand{\ioScoring}[4]{%
  \noindent{\color{IOBrown}\ioSmallCaps{Scored for} $u_g(t)$}\par
  \vspace{2pt}%
  \noindent\begingroup\setlength{\fboxsep}{3pt}%
  \colorbox{IOScorePaper}{\parbox[c][#1][c]{\dimexpr\linewidth-2\fboxsep\relax}{%
    \raggedright #3}}%
  \endgroup\par
  \vspace{2.5pt}%
  \noindent\parbox[t][#2][t]{\linewidth}{\raggedright #4}\par}
\newcommand{\ioPairRow}[2]{%
  \noindent\makebox[25pt][l]{\color{IOBrown}\ioSmallCaps{#1}}%
  \parbox[t]{\dimexpr\linewidth-25pt\relax}{\raggedright\ioCode{#2}\par}\par}
\newcommand{\ioOutcome}[6]{%
  \begin{minipage}[t]{\dimexpr(\linewidth-5pt)/2\relax}
    \raggedright\setlength{\fboxsep}{2.6pt}%
    \noindent\colorbox{#1}{\makebox[\dimexpr\linewidth-2\fboxsep\relax][l]{%
      \strut\color{IOBrown}\ioSmallCaps{#2}\hfill$D_t=#3$}}\par
    \vspace{2.2pt}%
    #4\par
    \vspace{2.2pt}%
    {\fontsize{7.6}{8.5}\selectfont\itshape\color{#5}#6\par}%
  \end{minipage}}
\newcommand{\ioCompare}[4]{%
  \par\vspace{4pt}\noindent
  \ioOutcome{IOCoral}{Disagree}{1}{#1}{IOCoralInk}{#2}\hfill
  \ioOutcome{IOAgree}{Agree}{0}{#3}{IOBrown}{#4}\par}
\floatname{algorithm}{Algorithm}
\renewcommand{\thealgorithm}{A\arabic{algorithm}}
\algrenewcommand{\algorithmicrequire}{\textbf{Input:}}
\algrenewcommand{\algorithmicensure}{\textbf{Output:}}
\algrenewcommand{\alglinenumber}[1]{\footnotesize\textcolor{black!60}{#1:}}
\algtext*{EndIf}
\algtext*{EndFor}
\definecolor{CompPhaseBand}{HTML}{F3F3F3}
\definecolor{CompPhaseNote}{HTML}{616161}
\newcommand{\CompPhase}[2]{%
  \par\addvspace{2.5pt}
  \Statex\hspace*{\algorithmicindent}%
  \colorbox{CompPhaseBand}{%
    \makebox[\dimexpr\linewidth-\algorithmicindent-2\fboxsep\relax][l]{%
      \strut\textbf{\textsc{#1}}\hfill\textcolor{CompPhaseNote}{\textit{#2}}}}}
\definecolor{NavMetricHeader}{HTML}{F5F5F6}
\providecolor{MainGray}{HTML}{F4F5F6}
\providecolor{MainAmber}{HTML}{FFF7E6}
\providecommand{\MainMetricWithDelta}[4]{%
  \mbox{\makebox[#1][r]{#3}\kern0.3pt%
    \makebox[#2][l]{%
      \raisebox{-0.65ex}{\fontsize{6}{6.4}\selectfont #4}%
    }%
  }%
}

\appendix
\raggedbottom
\section*{Appendix}
\numberwithin{equation}{section}
\setcounter{table}{0}
\renewcommand{\thetable}{A\arabic{table}}
\setcounter{figure}{0}
\renewcommand{\thefigure}{A\arabic{figure}}

Sec.~\ref{app:setup} provides experimental and deployment settings. Secs.~\ref{app:compensation}--\ref{app:online-control} develop the offline reconstruction, interface-specific calibration, and controller implementation of Sec.~\ref{sec:vlnq-method}; Secs.~\ref{app:draft-evaluation} and~\ref{app:qualitative-discussion} report recomputation measurements and additional navigation trajectories.

\section{Experimental and Deployment Details}
\label{app:setup}

\paragraph{Datasets and evaluation.}
The VLN-CE evaluation in Sec.~\ref{sec:experimental-setup} uses low-level navigation in scanned Matterport3D indoor environments through Habitat~\cite{krantz2020beyond,savva2019habitat}. R2R provides English room-to-room navigation instructions~\cite{anderson2018vision}, while RxR contains longer paths and more detailed instructions in English, Hindi, and Telugu~\cite{ku2020room}. Our evaluation uses English instructions in scenes disjoint from training.

\paragraph{Backbone implementation.}
We use the released checkpoints of the five configurations in Tab.~\ref{tab:main}, preserving their native image preprocessing, visual-history processing, and action execution. The quantization presets introduced in Sec.~\ref{sec:experimental-setup} specify the format of each language-backbone projection. Uni-NaVid uses FP16 language-backbone computation, while the other four configurations use BF16. The unquantized components follow Sec.~\ref{sec:vlnq-setup}. Quantized weights, correction factors, ranks, and thresholds are frozen before unseen-split evaluation.

\paragraph{Baseline implementation.}
The GPTQ-W4A16 baseline quantizes language-backbone linear projections, retaining the original precision of the visual encoder, multimodal projector, token embeddings, and output head. Its implementation uses packed weights with low-bit matrix--vector operations during decoding and per-projection BF16 reconstruction during prefill. Both GPTQ and AWQ follow the group-size-128 setting of Sec.~\ref{sec:experimental-setup}.

\paragraph{Deployment measurement.}
The GPTQ--ReCoNav comparison in Tab.~\ref{tab:main} uses RxR tasks completed successfully by the compared configurations. Full-precision and AWQ latency measurements use matched task episodes. Jetson timing spans RGB availability to action output, including visual processing, action generation, and triggered reruns. It excludes model loading, network transfer, simulation rendering, and time spent waiting for action execution. On the A800, we measure peak GPU memory for the inference process separately from Habitat rendering. Size counts deployed weights, quantization parameters, and compensation factors. Tab.~\ref{tab:scheduling} uses mixed R2R/RxR StreamVLN episodes, while Sec.~\ref{app:draft-evaluation} isolates compensated language-model recomputation on A800.

\section{Offline Compensation}
\label{app:compensation}

Offline preparation determines the correction factors and their allocation across projections by reconstructing quantization-induced projection-output residuals on recorded activations.

\subsection{Activation Collection}
The reconstruction trajectories introduced in Sec.~\ref{sec:vlnq-reconstruction} are driven by the full-precision model. At each recorded decision, the quantized branch receives the same observations and accepted interaction history, from which we collect its input activations $X_m$ at projection $m$. We retain at most 2048 activation rows per projection. Fitting and held-out reconstruction use disjoint training trajectories.

Factors are fitted separately for R2R and RxR; each model configuration uses a single control threshold across the two datasets (Sec.~\ref{app:calibration}).

\subsection{Activation-Weighted Low-Rank Reconstruction}

We solve the activation-weighted reconstruction objective in Eq.~\eqref{eq:vlnq-reconstruction} within the retained activation subspace of $X_m$. The weight residual $R_m$ and the correction factors follow the notation of Sec.~\ref{sec:vlnq-setup}.
\paragraph{Activation-weighted decomposition.}
We eigendecompose the sample Gram matrix $X_mX_m^{\mathsf T}=U_m\Lambda_mU_m^{\mathsf T}$, retaining at most $1024$ directions whose eigenvalues exceed $10^{-10}$ times the largest eigenvalue. Let $U_{m,k}$ and $\Lambda_{m,k}$ denote the retained eigenvectors and eigenvalues. We construct the activation basis as
\begin{equation}
 S_m=\Lambda_{m,k}^{1/2},\qquad
 V_m=X_m^{\mathsf T}U_{m,k}S_m^{-1}.
\label{eq:app-activation-basis}
\end{equation}
The quantization residual is projected onto this basis and weighted by the corresponding activation strengths:
\begin{equation}
 H_m=R_mV_mS_m.
\label{eq:app-weighted-residual}
\end{equation}
Let $Q_{m,r}$ contain the leading $r$ right singular vectors of $H_m$, obtained by eigendecomposing $H_m^{\mathsf T}H_m$. The correction factors are
\begin{equation}
 A_{m,r}=H_mQ_{m,r},\qquad
 B_{m,r}=V_mS_m^{-1}Q_{m,r}.
\label{eq:app-correction-factors}
\end{equation}
Under exact arithmetic, this construction yields a minimum-error rank-$r$ approximation within the retained activation subspace. All candidate ranks share the same ordered decomposition and use the corresponding leading columns of the factors.

\paragraph{Reconstruction error.}
All factors use 16-bit floating-point storage. StreamVLN, Uni-NaVid, and LightNav-0 store FP16 factors, with StreamVLN and LightNav-0 converting them to BF16 for execution. Both SimpleVLN configurations store and execute BF16 factors. We evaluate Eq.~\eqref{eq:vlnq-reconstruction} at deployment precision over the full set of fitting activations and use the resulting errors for rank allocation. This evaluation accounts for activation-subspace truncation and numerical precision; reconstruction quality is measured independently on held-out training trajectories.

\subsection{Budget-Constrained Rank Allocation}

For the allocation problem in Eq.~\eqref{eq:vlnq-rank-allocation}, each eligible projection $m\in\mathcal P$ admits a subset $\mathcal R_m\subseteq\{0,16,32,64,128\}$ of supported ranks. We write the byte cost of a rank-$r$ branch as $c_m(r)=2r(d_{\mathrm{in},m}+d_{\mathrm{out},m})$. The reference budget is $M=\sum_{m\in\mathcal P}c_m(64)$.

\paragraph{Rank allocation solver.}
We divide candidate storage costs by their greatest common divisor and solve the resulting multiple-choice knapsack problem by dynamic programming. Starting from $M$, we recover the minimum-fitting-error rank configuration under the current storage budget. If the mean rank exceeds 64, we tighten the budget by $0.02M$ and resolve the knapsack until both constraints in Eq.~\eqref{eq:vlnq-rank-allocation} are satisfied. This procedure approximately solves the joint storage- and mean-rank-constrained allocation.

\section{Action Scoring and Threshold Calibration}
\label{app:action-interfaces}

The controller in Sec.~\ref{sec:vlnq-scheduling} uses action-bearing output content to compute $u_g(t)$; offline calibration compares paired predictions to label $D_t$. We adapt these quantities to each backbone's native action representation, with the same decision input $x_t$ held fixed for the two compensation states. Fig.~\ref{fig:io-examples} illustrates scored positions and representative agreement and disagreement cases. The following rules specify the interface-dependent scores used in both calibration and online control.

\paragraph{Diagnostic in \cref{fig:uncertainty}.}
The base branch drives the training trajectory and the compensated branch predicts on the same history. StreamVLN and Uni-NaVid are shown on both datasets; the other three configurations are shown on R2R. Each panel bins decisions by base uncertainty and plots their disagreement rate. StreamVLN, Uni-NaVid, and LightNav-0 use ten bins from 50 training paths per model and dataset shown. Each SimpleVLN panel uses five bins from 16 paths. Shaded intervals are obtained by resampling paths. For SimpleVLN, this diagnostic scores action type and turn direction, and compares actions after merging consecutive actions of the same type, ignoring magnitude-only differences. The deployment score and disagreement label in Sec.~\ref{app:text-commands} additionally account for movement magnitude.

\subsection{Discrete Actions}

StreamVLN represents movement using direction symbols, while Uni-NaVid generates action words. Both use the legal action set $\mathcal A=\{\texttt{stop},\texttt{forward},\texttt{left},\texttt{right}\}$. Let $a_{t,j}^g$ denote the generated action at a valid scored position $j\in \mathcal J_t^g$, and $z_{t,j}^g(v)$ the score of candidate action $v$. The action probability in Eq.~\eqref{eq:vlnq-uncertainty} is
\begin{equation}
 p_{t,j}^g = \frac{\exp z_{t,j}^g(a_{t,j}^g)}{\sum_{v\in\mathcal A}\exp z_{t,j}^g(v)}.
\end{equation}
We exclude formatting tokens and average over valid positions as in Eq.~\eqref{eq:vlnq-uncertainty}. For disagreement, the prefixes in Eq.~\eqref{eq:vlnq-disagreement} are obtained as $\tilde a_t^{(g)}=\Pi_b(y_t^{(g)})$, where $\Pi_b$ reads the first four entries according to backbone $b$'s native action-output convention. Fig.~\ref{fig:io-examples} (top) illustrates how action identity and order affect the label.

\subsection{Spatial Pointers}
\label{app:spatial-pointers}

LightNav-0 generates spatial pointers and RVQ motion codes. The \texttt{apos} and \texttt{opos} pointers locate an affordance and an object on an image grid, respectively, while RVQ codes are decoded into a motion trajectory through the model's native interface~\cite{wang2026lightnav}. Spatial candidates have a geometric structure: probability spread over neighboring cells need not indicate spatial ambiguity. We therefore measure confidence through local probability mass rather than the generated position's probability alone. For each pointer, we sum the mass among the five highest-scoring candidate positions that lies within radius one of the generated position.

For pointer type $\rho\in\{\mathrm a,\mathrm o\}$, let $z_{t,\rho}^g(c)$ denote the generation score of position $c$, $\mathcal K_{t,\rho}^g$ the five highest-scoring positions, and $\hat c_{t,\rho}^g$ the generated position. We compute
\begin{equation}
\begin{aligned}
 p_{t,\rho}^g&=\frac{\sum_{\substack{c\in\mathcal K_{t,\rho}^g\\d(c,\hat c_{t,\rho}^g)\le1}}\exp z_{t,\rho}^g(c)}
 {\sum_{c\in\mathcal K_{t,\rho}^g}\exp z_{t,\rho}^g(c)},\\
 u_g(t)&=-\tfrac12\sum_{\rho\in\{\mathrm a,\mathrm o\}}\log p_{t,\rho}^g.
\end{aligned}
\label{eq:app-lightnav-uncertainty}
\end{equation}
Ordinary position IDs $1\le c\le1296$ map to a $48$-column grid with coordinates $(\lfloor(c-1)/48\rfloor,(c-1)\bmod48)$, where $d$ is Euclidean grid distance. A special position token has distance zero to itself and infinity to other tokens. A valid score requires exactly one pointer of each type, with each generated position among its type's five highest-scoring positions.

For paired predictions, agreement tolerates a one-cell shift along either axis, including diagonal neighbors. Thus, using the same grid distance $d$, we define
\begin{equation}
 D_t=\mathbf1\!\left\{\exists\rho,\ d(\hat c_{t,\rho}^{0},\hat c_{t,\rho}^{1})>\sqrt{2}\right\}.
\label{eq:app-lightnav-disagreement}
\end{equation}
Pairs missing either pointer type receive no disagreement label. The $\sqrt{2}$ tolerance in the comparison includes diagonal neighbors, whereas the uncertainty score uses radius one. This label measures changes in the spatial pointers; subsequent RVQ motion codes are outside its scope (Fig.~\ref{fig:io-examples}, bottom left).

\subsection{Text Commands}
\label{app:text-commands}

SimpleVLN produces text commands with semantic fields for action type, turn direction where applicable, and movement magnitude~\cite{hong2026simplevln}. Both model sizes score the fields of the first two commands following the reasoning segment. Legal values are \texttt{stop}/\texttt{forward}/\texttt{turn} for action type, \texttt{left}/\texttt{right} for direction, 25/50/75 cm for distance, and $15^\circ/30^\circ/45^\circ$ for turn angle.

Let $\mathcal V_f$ contain all legal tokens for field $f$, and $\mathcal V_f(v_{t,f}^g)$ the tokens corresponding to the generated field value. Using token scores $z_{t,f}^g(i)$ at the field's first position, we compute
\begin{equation}
\begin{aligned}
 p_{t,f}^g&=\frac{\sum_{i\in\mathcal V_f(v_{t,f}^g)}\exp z_{t,f}^g(i)}
 {\sum_{i\in\mathcal V_f}\exp z_{t,f}^g(i)},\\
 u_g(t)&=-\frac{1}{|\mathcal F_t^g|}\sum_{f\in\mathcal F_t^g}\log p_{t,f}^g.
\end{aligned}
\label{eq:app-simplevl-uncertainty}
\end{equation}
Here, $\mathcal F_t^g$ contains successfully aligned and scored fields. Candidate sets include capitalization variants of each value. Numerical fields are scored using the first numerical token. Reasoning text, punctuation, units, and end tokens are excluded from the average.

Disagreement is evaluated in the space of canonical primitive actions. Let $C(y)$ expand the first two commands into ordered forward steps of 25 cm and turns of $15^\circ$, with at most three repetitions per command. A stop command terminates the sequence. We define
\begin{equation}
 D_t=\mathbf1\!\left\{C(y_t^{(0)})\ne C(y_t^{(1)})\right\}.
\label{eq:app-simplevl-disagreement}
\end{equation}
For example, \texttt{forward 50 cm} and two successive \texttt{forward 25 cm} commands both yield two forward steps and therefore agree (Fig.~\ref{fig:io-examples}, bottom right). Pairs with an empty sequence or an unparsed-command marker receive no disagreement label.

\subsection{Paired Calibration and a Shared Threshold}
\label{app:calibration}

\paragraph{Paired data collection.}
For each dataset, we select 50 distinct training paths. Each model configuration runs every path in two independent rollouts, driven by the base and compensated branches, respectively. At each decision, both branches predict from the same input $x_t$, yielding $u_0(t)$, $u_1(t)$, and $D_t$. Only the driving branch advances the environment and updates the accepted history; the other supplies the paired prediction. This protocol samples decision contexts under both operating states.

\paragraph{Shared-threshold selection.}
For each calibration fit, we pool the valid base and compensated uncertainty scores from its training records and sort the distinct values. The candidate set $\mathcal T_b$ comprises the midpoints between consecutive scores, a boundary slightly below the minimum, and the maximum score, covering all distinct partitions under the strict comparison $u>\tau$.

For each candidate, we replay the state transitions in Eqs.~\eqref{eq:vlnq-accept} and~\eqref{eq:vlnq-next} sequentially over the paired records, resetting the compensation state to off at the start of each rollout. Records without a valid disagreement label remain in the replay sequence but do not contribute to disagreement F1. The recorded inputs and predictions remain fixed during replay. Disagreement and accepted-compensation labels follow Sec.~\ref{sec:vlnq-scheduling}. We select the threshold using the equally weighted R2R/RxR disagreement-F1 objective in Eq.~\eqref{eq:vlnq-calibration}.

\paragraph{Path-grouped five-fold evaluation.}
For each dataset, we partition the 50 paths into five groups of ten, keeping all decisions from both rollouts of a path in the same fold. In each fold, we recalibrate the shared threshold on 40 paths per dataset and evaluate it on the remaining ten paths per dataset. The held-out paired records are replayed in temporal order using the same initialization and state-transition rules. We compute the equally weighted mean of the two dataset-level disagreement F1 scores for each fold.

For deployment, each model configuration's threshold is fitted on all paired training records using Eq.~\eqref{eq:vlnq-calibration} and frozen before unseen-split evaluation.

\begin{figure*}[!t]
\centering
\begin{minipage}{\linewidth}
\centering
\begin{tcbraster}[raster columns=2,raster equal height=rows,
  raster column skip=0.03\textwidth,raster row skip=6pt]
\begin{nativecard}{StreamVLN}{Direction symbols}
  \ioScoring{13pt}{11pt}{\ioScored{$\uparrow\quad\uparrow\quad\leftarrow\quad\uparrow$}
    \enspace\ioCode{\ioIgnored{\iotoken{<|im_end|>}}}}%
    {Score the ordered action symbols.}
  \ioCompare{%
    \ioPairRow{Base}{$\uparrow\quad\uparrow\quad\leftarrow\quad\uparrow$}
    \ioPairRow{Comp.}{$\uparrow\quad\textcolor{IOCoralInk}{\boldsymbol{\rightarrow\quad\uparrow\quad\leftarrow}}$}%
  }{Different action order.}{%
    \ioPairRow{Base}{$\rightarrow\quad\uparrow\quad\rightarrow\quad\uparrow$}
    \ioPairRow{Comp.}{$\rightarrow\quad\uparrow\quad\rightarrow\quad\uparrow$}%
  }{Same four-action prefix.}
\end{nativecard}
\begin{nativecard}{Uni-NaVid}{Action words}
  \ioScoring{13pt}{11pt}{\ioScored{\ioCode{\texttt{right forward right forward}}}
    \enspace\ioCode{\ioIgnored{\iotoken{</s>}}}}%
    {Score action words within the first four generated positions.}
  \ioCompare{%
    \ioPairRow{Base}{\texttt{right forward}\newline\texttt{right forward}}
    \ioPairRow{Comp.}{\ioChanged{\texttt{left}}\texttt{ forward}\newline\texttt{right forward}}%
  }{The first word changes.}{%
    \ioPairRow{Base}{\texttt{right forward}\newline\texttt{right forward}}
    \ioPairRow{Comp.}{\texttt{right forward}\newline\texttt{right forward}}%
  }{All four words match.}
\end{nativecard}
\begin{nativecard}{LightNav-0}{Spatial pointers}
  \ioScoring{27pt}{20pt}{\ioScored{\ioCode{\iotoken{<apos_980>}}}\,
    \ioScored{\ioCode{\iotoken{<opos_884>}}}
    \enspace\ioIgnored{Base cells $(20,19)$ and $(18,19)$}}%
    {Score the top-5 probability mass within radius one for each pointer (affordance and object).}
  \ioCompare{%
    \ioPairRow{Comp.}{\ioChanged{\iotoken{<apos_982>}}\newline\ioChanged{\iotoken{<opos_886>}}}%
  }{Two cells, $d=2>\sqrt{2}$.}{%
    \ioPairRow{Comp.}{\iotoken{<apos_1029>}\newline\iotoken{<opos_933>}}%
  }{Diagonal neighbor, $d=\sqrt{2}$.}
\end{nativecard}
\begin{nativecard}{SimpleVLN}{Text commands}
  \ioScoring{27pt}{20pt}{\ioCode{%
    \ioIgnored{\iotoken{</think>}}\par
    \ioScored{\texttt{forward 50}}\ioIgnored{\texttt{ cm, }}%
    \ioScored{\texttt{turn left 15}}\ioIgnored{\texttt{ degree,}}\par
    \ioIgnored{\texttt{forward 25 cm}\iotoken{<|im_end|>}}}}%
    {Score action, direction, and numeric fields in the first two commands. Use the first token of each number.}
  \ioCompare{%
    \ioPairRow{Base}{\texttt{forward 50 cm}}
    \ioPairRow{Comp.}{\texttt{forward }\ioChanged{\texttt{25}}\texttt{ cm}}%
  }{Two forward steps vs. one.}{%
    \ioPairRow{Base}{\texttt{forward 50 cm}}
    \ioPairRow{Comp.}{\texttt{forward 25 cm,}\newline\texttt{forward 25 cm}}%
  }{Both give two $25$ cm steps.}
\end{nativecard}
\end{tcbraster}
\captionof{figure}{\textbf{Uncertainty scoring and action disagreement across navigation interfaces.}
Amber highlights the output positions used to compute $u_g(t)$.
The lower panels compare outputs for the same input with compensation off (Base) and on (Comp.).
Disagreement $D_t$ is used to calibrate the compensation threshold.}
\label{fig:io-examples}
\end{minipage}
\end{figure*}

\section{Online Control and Draft Verification}
\label{app:online-control}

\subsection{Controller Implementation}
Algorithm~\ref{alg:app-compensation} implements the state transitions in Eqs.~\eqref{eq:vlnq-accept} and~\eqref{eq:vlnq-next}. The compensation state is reset to off at the start of each rollout. The controller applies an uncertainty test only when its score is available: an unavailable base score triggers no rerun, and an unavailable compensated score retains compensation for the next decision. Calibration replay uses the same fallback. Only the accepted output is executed and added to the interaction history.

\begin{algorithm}[!t]
\caption{Action-uncertainty-driven compensation}
\label{alg:app-compensation}
\fontsize{9}{9.7}\selectfont
\setlength{\fboxsep}{1.7pt}
\begin{algorithmic}[1]
\Require Frozen weights, factors, threshold $\tau$
\Ensure Executed actions and updated interaction history
\State $s_1\gets0$
\For{each decision $t$}
    \CompPhase{Generate a candidate}{current state}
    \State Prepare $x_t$; save language state
    \State $g_t\gets s_t$
    \State $(y_t^{(g_t)},u_{g_t}(t))\gets$ \Call{Generate}{$x_t,g_t$}
    \CompPhase{Verify the draft}{on trigger}
    \If{$s_t=0\land$ \Call{Valid}{$u_0(t)$} $\land u_0(t)>\tau$}
        \State Restore state; reuse visual features
        \State $g_t\gets1$
        \State $(y_t^{(1)},u_1(t))\gets$ \Call{VerifyDraft}{$x_t,y_t^{(0)}$}
    \EndIf
    \CompPhase{Execute and update}{next decision}
    \State \Call{Execute}{$y_t^{(g_t)}$}; update $h_{t+1}$
    \State $s_{t+1}\gets g_t$
    \If{$g_t=1\land$ \Call{Valid}{$u_1(t)$}}
        \State $s_{t+1}\gets\mathbf1\{u_1(t)>\tau\}$
    \EndIf
\EndFor
\end{algorithmic}
\end{algorithm}

\subsection{State Restoration and Draft Verification}
\label{app:implementation}

For the rerun in Sec.~\ref{sec:vlnq-scheduling}, we restore the pre-decision language state and cache while reusing the visual features computed by the base pass. This prevents duplicate visual-history updates and supplies the compensated branch with the same decision input.

The base output $y_t^{(0)}$ serves as the verification draft. A compensated forward pass over draft tokens scores multiple positions together; the backbone's decoding rule determines the longest matching prefix. At the first mismatch, the compensated token replaces the draft token and autoregressive decoding resumes. Under matched initial states, deterministic decoding, and consistent logits, this procedure preserves the output of standard compensated autoregressive decoding from the same state.

For attention-based states, we crop key--value caches to the accepted position; for recurrent states, we recompute the accepted prefix to restore the cache. We compute $u_1(t)$ from compensated generation scores, including the scores of accepted draft tokens. The completed compensated output determines the executed action and the continuation state through Eq.~\eqref{eq:vlnq-next}. Sec.~\ref{app:draft-evaluation} quantifies the savings from this procedure.

\section{Draft-Verification Microbenchmark}
\label{app:draft-evaluation}

\paragraph{Metrics and timing protocol.}
To isolate the cost of the rerun procedure in Sec.~\ref{app:implementation}, we collect compensated-rerun inputs from 25 randomly selected R2R trajectories and 25 randomly selected RxR trajectories and replay them on an NVIDIA A800 GPU. We compare draft verification with standard compensated autoregressive decoding under identical compensation factors and decoding settings. Acceptance is the fraction of verified draft tokens retained by the compensated branch. Forward calls are averaged per rerun and include prefill, verification, autoregressive continuation, and cache reconstruction where applicable.

\begin{table}[tbp]
\captionsetup{hypcap=false,skip=6pt,justification=raggedright,singlelinecheck=false}
\caption{\textbf{Draft verification for compensated language-model reruns on A800.} AR denotes standard autoregressive decoding. Calls are averages per rerun; speedup is computed from the aggregate median times in Eq.~\eqref{eq:app-draft-speedup}.}
\label{tab:app-draft}
\centering
\fontsize{8.8}{10.6}\selectfont
\setlength{\tabcolsep}{2.75pt}
\renewcommand{\arraystretch}{1.15}
\begin{tabularx}{\linewidth}{>{\raggedright\arraybackslash}Xrrrr}
\toprule
\rowcolor{MainGray}
\textbf{Backbone} & \textbf{Acceptance} & \textbf{AR calls} & \textbf{Draft calls} & \textbf{Speedup} \\
\midrule
StreamVLN & $84.38\%$ & $8.00$ & \cellcolor{MainAmber}$1.90$ & \cellcolor{MainAmber}{\boldmath $3.29\times$} \\
Uni-NaVid & $62.68\%$ & $5.00$ & \cellcolor{MainAmber}$2.39$ & \cellcolor{MainAmber}{\boldmath $1.69\times$} \\
LightNav-0 & $42.33\%$ & $6.00$ & \cellcolor{MainAmber}$3.76$ & \cellcolor{MainAmber}{\boldmath $1.49\times$} \\
SimpleVLN-4B & $71.78\%$ & $18.00$ & \cellcolor{MainAmber}$5.31$ & \cellcolor{MainAmber}{\boldmath $1.64\times$} \\
SimpleVLN-9B & $77.30\%$ & $20.34$ & \cellcolor{MainAmber}$4.95$ & \cellcolor{MainAmber}{\boldmath $1.74\times$} \\
\bottomrule
\end{tabularx}
\end{table}

For each rerun input $i$, we alternate repeated measurements of both methods and obtain their median times, $T_{\mathrm{AR},i}$ and $T_{\mathrm{draft},i}$. Aggregate speedup is
\begin{equation}
\mathrm{Speedup}=
\frac{\sum_i T_{\mathrm{AR},i}}{\sum_i T_{\mathrm{draft},i}}.
\label{eq:app-draft-speedup}
\end{equation}
Timing covers compensated language-model recomputation only, excluding visual encoding, the initial base generation, model loading, and environment interaction.

\paragraph{Recomputation efficiency.}
Draft verification accelerates compensated language-model reruns by 1.49--3.29$\times$ across the five configurations (Tab.~\ref{tab:app-draft}). StreamVLN retains 84.38\% of verified draft tokens and reduces the mean forward-call count from 8.00 to 1.90, yielding a 3.29$\times$ speedup. These results quantify the savings per rerun, complementing the reduction in rerun frequency reported in Sec.~\ref{sec:allocation}.

\section{Qualitative Navigation Analysis}
\label{app:qualitative-discussion}

\begin{figure*}[p]
  \centering
  \includegraphics[
    width=\textwidth,
    height=\dimexpr\textheight-25pt\relax,
    keepaspectratio
  ]{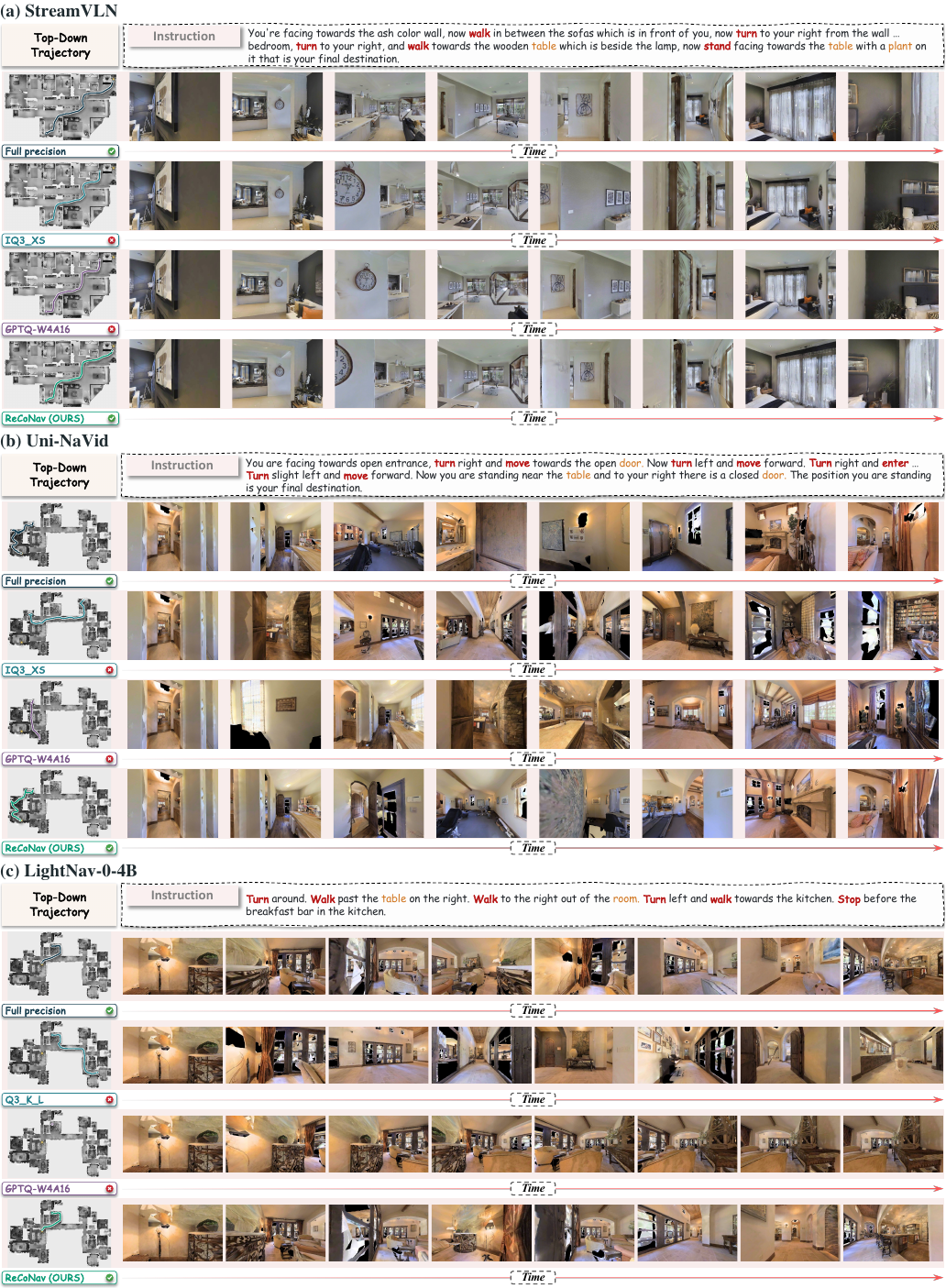}
\caption{\textbf{Selected navigation trajectories for StreamVLN, Uni-NaVid, and LightNav-0.} Each episode compares four deployments using top-down paths and temporally ordered observations.}
  \label{fig:app-qualitative-first}
\end{figure*}

\begin{figure*}[p]
  \centering
  \includegraphics[width=\textwidth,height=\dimexpr\textheight-25pt\relax,keepaspectratio]{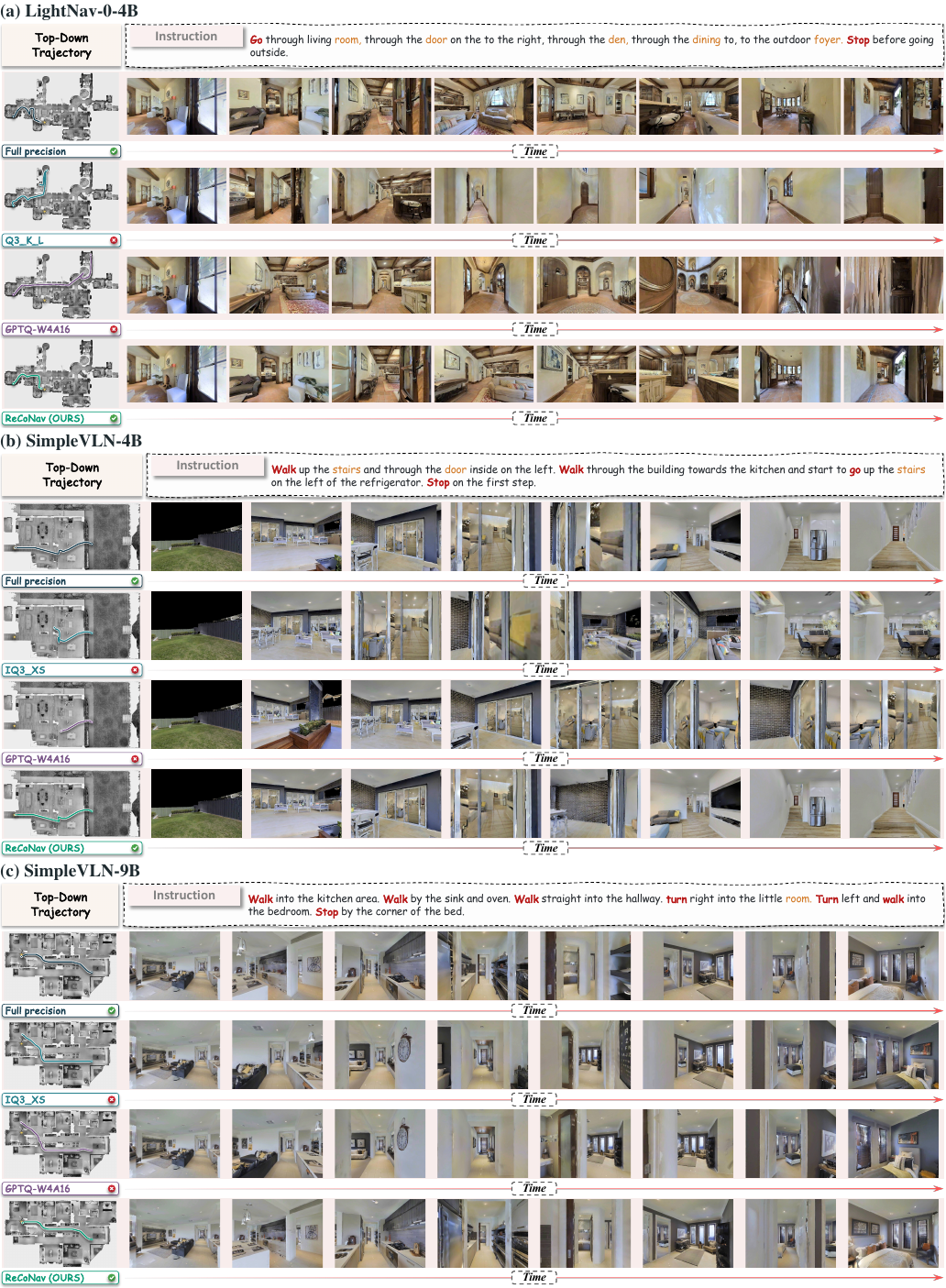}
\caption{\textbf{Selected navigation trajectories for LightNav-0 and SimpleVLN-4B/9B.} Each episode uses the same four-deployment visualization protocol as Fig.~\ref{fig:app-qualitative-first}.}
  \label{fig:app-qualitative-second}
\end{figure*}

Figs.~\ref{fig:app-qualitative-first} and~\ref{fig:app-qualitative-second} extend the qualitative comparisons in Sec.~\ref{sec:main-results} to all five model configurations, including two examples for LightNav-0. Each selected episode compares full precision, uncompensated quantization, GPTQ-W4A16, and ReCoNav. The top-down paths show the executed trajectories, accompanied by eight observations sampled in temporal order. Uncompensated baselines use \texttt{IQ3\_XS}, except LightNav-0, which uses \texttt{Q3\_K\_L}.

In all six selected episodes, full precision and ReCoNav reach the goal, whereas uncompensated quantization and GPTQ-W4A16 fail.

\FloatBarrier

\clearpage

\end{document}